\documentclass[accepted]{uai2026} 

\usepackage[american]{babel}

\usepackage{natbib} 
\usepackage{mathtools} 
\usepackage{booktabs} 
\usepackage{tikz} 

\usepackage{titletoc}
\usepackage{graphicx}
\usepackage{url}            
\usepackage{booktabs}       
\usepackage{amsfonts}       
\usepackage{nicefrac}       
\usepackage{microtype}      
\usepackage{xcolor}         

\usepackage{amsmath}
\usepackage{amssymb}
\usepackage{mathtools}
\usepackage{amsthm}
\usepackage{hyperref}

\usepackage{algorithm}
\usepackage{algorithmic}

\usepackage{amsmath}
\usepackage{amssymb}
\usepackage{mathtools}
\usepackage{amsthm}
\usepackage{hyperref}

\usepackage{amsmath,amsfonts,bm,bbm}

\newcommand{\Exp}{\operatorname{Exp}}

\usepackage{cleveref}
\usepackage{thmtools}
\newtheorem{theorem}{Theorem}
\newtheorem{lemma}[theorem]{Lemma}

\newtheorem{definition}[theorem]{Definition}
\newtheorem{assumption}[theorem]{Assumption}

\Crefname{ALC@line}{Line}{Lines}
\Crefname{assumption}{Assumption}{Assumptions}
\Crefformat{equation}{Eq. #2(#1)#3}
\Crefrangeformat{equation}{Eqs. #3(#1)#4 to #5(#2)#6}
\Crefmultiformat{equation}{Eqs. #2(#1)#3}{ and #2(#1)#3}{, #2(#1)#3}{ and #2(#1)#3}
\Crefrangemultiformat{equation}{Eqs. #3(#1)#4 to #5(#2)#6}{ and #3(#1)#4 to #5(#2)#6}{, #3(#1)#4 to #5(#2)#6}{ and #3(#1)#4 to #5(#2)#6}

\def\eqref#1{equation~\ref{#1}}

\def\ceil#1{\lceil #1 \rceil}

\def\1{\mathbbm{1}}

\def\rd{{\textnormal{d}}}

\def\vp{{\bm{p}}}
\def\vq{{\bm{q}}}

\def\vx{{\bm{x}}}

\def\mS{{\bm{S}}}

\DeclareMathAlphabet{\mathsfit}{\encodingdefault}{\sfdefault}{m}{sl}
\SetMathAlphabet{\mathsfit}{bold}{\encodingdefault}{\sfdefault}{bx}{n}

\def\gB{{\mathcal{B}}}

\def\gE{{\mathcal{E}}}
\def\gF{{\mathcal{F}}}

\def\gH{{\mathcal{H}}}

\def\gL{{\mathcal{L}}}

\def\gO{{\mathcal{O}}}

\def\gR{{\mathcal{R}}}
\def\gS{{\mathcal{S}}}

\def\Prob{{\mathbb{P}}}

\newcommand{\wt}{\widetilde}

\newcommand{\E}{\mathbb{E}}

\newcommand{\R}{\mathbb{R}}

\newcommand{\KL}{D_{\mathrm{KL}}}

\newcommand{\EE}{\mathbb{E}}

\newcommand{\cO}{\mathcal{O}}

\newif\ifsup\supfalse
\suptrue

\DeclareMathOperator*{\argmax}{argmax}
\DeclareMathOperator*{\argmin}{argmin}

\title{On the Sublinear Regret of Continuous K-Max Bandits}

\author[1]{Yu~Chen\thanks{Equal contribution. This work was done while Yu Chen was an intern at Microsoft Research Asia. Corresponding author: Wei Chen (weic@microsoft.com).}}
\author[2]{Siwei~Wang$^*$}
\author[1]{Longbo~Huang}
\author[2]{Wei~Chen}
\affil[1]{%
    Institute for Interdisciplinary Information Sciences\\
    Tsinghua University
}
\affil[2]{%
    Microsoft Research Asia
}

\begin{document}
\maketitle

\begin{abstract}
The $K$-Max combinatorial multi-armed bandit problem arises in applications such as recommendation and distributed decision making, where the reward is determined by the maximum outcome among $K$ selected arms. When outcomes are continuous and only the maximum value together with the winner's index is observed, this problem introduces unprecedented difficulties including discretization errors, non-deterministic tie-breaking, and severe estimation biases. To overcome these barriers, we introduce \texttt{DCK-UCB}, an efficient algorithm combining adaptive discretization with bias-corrected confidence bounds. We prove that \texttt{DCK-UCB} achieves a $\widetilde{\mathcal{O}}(T^{3/4})$ regret bound, the \emph{first} sublinear guarantee in this setting. Numerical experiments show strong performance over baseline methods. Furthermore, for the specific case of exponential distributions under full-bandit feedback, we propose the \texttt{MLE-Exp} algorithm that attains a near-optimal $\widetilde{\mathcal{O}}(\sqrt{T})$ regret bound. This work establishes fundamental theoretical guarantees and provides a powerful algorithmic solution for continuous combinatorial bandits.
\end{abstract}

\section{Introduction}

Combinatorial MABs (CMABs) \citep{cesa2012combinatorial,chen2013combinatorial} have gained significant attention due to applications in online advertising, networking, and influence maximization \citep{gai2012combinatorial,kveton2015combinatorial, chen2009efficient,chen2013combinatorial}. In CMABs, an agent selects a subset of arms as the combinatorial action for each round, and the environment returns a reward signal according to the outcomes of the selected arms.
As a popular variant,  \emph{$K$-Max Bandits} \citep{goel2006asking} focuses on the maximum outcome within a selected subset of $K$ arms, naturally capturing real-world scenarios where the decision quality depends only on the best realized outcome.

In this paper, we consider the \textit{Continuous} $K$-Max Bandits problem where the outcome for each arm follows a continuous distribution, motivated by the prevalence of continuous real-valued signals (e.g., job completion time in distributed computing, latency in server scheduling, bidding price in auctions, and ratings of selected products in online advertising).
Moreover, we focus on a limited feedback environment that provides only the maximum outcome (reward) and the index of the winning arm (the arm that achieved the maximum).  This novel framework, termed \emph{Continuous $K$-Max Bandits} with \emph{value-index feedback}, is applicable to various real-world scenarios.
For example,
		in {distributed computing tasks}, a scheduler may choose $K$ servers for parallel processing and the overall completion metric depends on the server with the fastest response while feedback from others can be overshadowed. The latency we need to optimize is the shortest response time among the selected servers.
Likewise,
the seller invites a fixed number of bidders to a first-price auction; each invited bidder independently draws and submits a bid from an unknown continuous valuation distribution; the highest bid wins and is paid, and the seller observes only the winning bid and the winner’s identity, while all other bids remain unobserved.

Tackling this problem is highly challenging.
Most existing studies for $K$-Max bandits focus on discrete outcome settings \citep{simchowitz2016best,wang2023combinatorial} or require feedback on every selected arm (semi-bandit feedback) \citep{chen2016combinatorial, wang2017improving}. For the continuous $K$-Max bandits, naively applying existing discrete approaches can lead to significant discretization errors and impractical operations, hindering learning efficiency. Furthermore, the combination of continuous outcomes and limited \emph{value-index feedback}
introduces unique challenges from biased observations, which are not typically faced in discrete settings or with richer feedback.

Specifically, existing studies
face three critical limitations when tackling the continuous distribution and value-index feedback:
First, most algorithms require semi-bandit feedback \citep{chen2016combinatorial,simchowitz2016best,wang2017improving} revealing all selected arms' outcomes, while practical systems often only provide the winning arm's index and value, introducing bias since observations occur only when an arm wins.
Second, greedy-style submodular bandit approaches \citep{streeter2008online,fourati2024combinatorial} typically compete with a $(1-1/e)$-approximation benchmark \citep{nemhauser1978analysis}, which can result in linear regret under our exact-optimum benchmark.
Third, most prior solutions for $K$-Max bandits assume binary \citep{simchowitz2016best} or
finitely supported outcomes \citep{wang2023combinatorial}, struggling with continuous distributions under value-index feedback due to: (i) discretization error versus learning efficiency trade-offs, (ii) biased estimation from limited observability, and (iii) non-deterministic tie-breaking when several continuous values map to one discrete bin. We further analyze the difficulties and challenges in \Cref{sec:challenges}.

To fully address these issues, we design a novel framework for Continuous $K$-Max Bandits with value-index feedback through two key technical innovations:
\textbf{(i)} We develop a principled approach to discretize the continuous outcome distributions, transforming the problem into a discrete $K$-Max bandit instance (\Cref{sec:discretization}). This allows us to control discretization error and utilize an efficient offline $(1-\epsilon)$-approximated optimization oracle (for any $\epsilon\in(0,1)$),
crucially avoiding the linear regret associated with traditional greedy algorithms limited by $(1-1/e)$ approximations.
\textbf{(ii)} Due to the nondeterministic tie-breaking effect arising from the continuous-to-discrete transformation under value-index feedback, the agent cannot achieve an unbiased estimation under computationally tractable discretization (\Cref{sec:algorithm}). In this paper, we design a bias-corrected estimation method coupled with a novel concentration analysis incorporating bias-aware error control (\Cref{sec:bias-corrected-estimators}), enabling us to jointly manage estimation variance and discretization-induced bias to achieve sublinear regret.

\textbf{Contributions.} Our contribution can be summarized as follows:

\textbf{(i)} Our primary contribution is algorithm \texttt{DCK-UCB} (\Cref{alg} in \Cref{sec:kmax}), designed for $K$-Max bandits with general continuous distributions under value-index feedback. We prove that \texttt{DCK-UCB} achieves a regret upper bound of $\widetilde{\mathcal{O}}(T^{3/4})$ (\Cref{thm:main}), which is the first polynomial-time algorithm to achieve a sublinear regret guarantee for continuous $K$-Max bandits with value-index feedback. We also provide numerical experiments in Appendix~\ref{app:experiments} to empirically validate the effectiveness of \texttt{DCK-UCB}.

\textbf{(ii)} The most interesting technical innovation is the newly developed bias-corrected estimation in \texttt{DCK-UCB}. Specifically, this technique helps us conduct efficient learning under biased observations due to the nondeterministic tie-breaking arising from the discretization.
This novel technique can be of independent interest.

\textbf{(iii)} We further develop the Maximum Likelihood Estimation (MLE) under a special case where outcomes for each arm follow exponential distributions. Our algorithm, \texttt{MLE-Exp} (\Cref{alg:k-min}), designed for $K$-Min bandits (a variant of continuous $K$-Max), leverages the unique structure of exponential distributions to bypass the need for discretization and its associated biases.
Combined with the lower bound $\Omega(\sqrt{T})$ proven in Appendix~\ref{app:lowerbound},  \texttt{MLE-Exp} achieves a near-optimal $\wt{\gO}(\sqrt{T})$ regret (\Cref{thm:kminexp}), representing a significant improvement over the general continuous case.

\textbf{Notations.}
For any integer $n \ge 1$, we use $[n]$ to denote the set $\{1,2,\ldots,n\}$. $\gO$ is used to suppress all constant factors, while $\wt{\gO}$ further suppresses all logarithmic factors. Bold letters such as $\vx$ are typically used to represent a set of elements $\{x_i\}$. Unless expressly stated, $\log(x)$ refers to the natural logarithm of $x$. Throughout the text, $\{\mathcal F_t\}_{t=0}^T$ is used to denote the natural filtration; that is, $\mathcal F_t$ represents the $\sigma$-algebra generated by all random observations made within the first $t$ time slots.

\section{Related Works}\label{sec:related-works}
The $K$-Max bandit problem departs from standard Combinatorial Multi-Armed Bandits (CMABs) \citep{cesa2012combinatorial,chen2013combinatorial,chen2014combinatorial,kveton2015tight,combes2015combinatorial,liu2023contextual},
	which suffices to learn the expected outcome of each arm.
In contrast, $K$-Max bandits, whose reward is the maximum outcome among the selected arms \citep{goel2006asking,gopalan2014thompson}, require knowing full distributions.
In the following, we organize related work by feedback type.

\textbf{Full-Bandit (Value) Feedback.}
Here the learner observes only the numerical reward, i.e., the maximum outcome over the chosen subset. \citet{gopalan2014thompson} derived a Thompson Sampling regret bound of $\mathcal{O}\!\left(\sqrt{\binom{N}{K}T}\right)$, but their analysis assumes a known finite parameter set and the regret scales exponentially in $K$. For pure exploration, \citet{simchowitz2016best} study the Bernoulli case. A submodular-maximization view yields $(1-1/e)$ approximation guarantees via greedy selection \citep{streeter2008online,yue2011linear,nie2022explore,fourati2024combinatorial}; $K$-Max naturally satisfies submodularity. However, such approximation-style guarantees lead to linear regret when the baseline is the \emph{true} optimal subset.
Achieving sublinear regret for general $K$-Max with full-bandit feedback therefore remains open.

\textbf{Value–Index Feedback.}
The learner observes both the maximum outcome and the index of the winning arm. This resembles CMAB with probabilistic triggering \citep{wang2017improving,liu2023contextual,liu2024combinatorial}.
Under this feedback model, \citet{wang2023combinatorial} analyze discrete $K$-Max with finite outcome support and deterministic tie-breaking and obtain $\wt{\gO}(\sqrt{T})$ regret; \citet{simchowitz2016best} also study Bernoulli outcomes.
However, these finite-support methods do not extend to the continuous setting considered in this paper due to non-zero discretization error and nondeterministic tie-breaking. DART \citep{agarwal2020dart} is another related value-index method that avoids finite-support enumeration via scalar arm-level estimates, but it relies on a structural ordering condition, informally that ``good arms generate good actions'': replacing a worse arm by a better one should improve the expected joint reward in every fixed context. Such a context-independent total order need not exist for general $K$-Max rewards, because $r^*(S)=\E[\max_{i\in S}X_i]$ depends on the full outcome distributions and the relative value of two arms can flip with the other arms in the selected set.

\textbf{Semi-Bandit Feedback.}
Semi-bandit reveals the outcomes of \emph{all} selected arms, enabling unbiased per-arm estimation and simpler UCB/TS-style learning. This richer feedback supports $\mathcal{O}(\sqrt{T})$ regret for discrete or continuous $K$-Max settings \citep{simchowitz2016best,jun2016top,chen2016combinatorial,chen2016combinatorialt,slivkins2019introduction}. Nonetheless, these algorithms rely on observations strictly richer than value-only or value–index feedback and are not directly applicable to our setting.

\section{Formulation}\label{sec:preliminaries}

We study the \textit{continuous $K$-Max bandits}, denoted as $\mathcal{B}^*$, where an agent interacts with $N$ arms $\mathcal{A} = [N]$. For each arm $i \in [N]$, there is a continuous random distribution $D_i$ such that $X_i \sim D_i$, where $X_i$ is the outcome of arm $i$.

The agent will play $T$ rounds in total.
At each time step $t \in [T]$, the agent needs to select an action $S_t$ from the feasible action set $\mathcal{S}=\{S\subseteq\mathcal{A}\mid|S|=K\}$, i.e., a subset of $\mathcal{A}$ with size $K$\footnote{Indeed, our theoretical guarantees and algorithmic approach naturally extend to all subsets $S$ with $|S| \le K$. When the super arm set is not the set of all subsets of cardinality at most $K$, our approach still works as long as there exists an efficient oracle to solve the offline optimization problem.}. Here $1 < K < N$ is a given constant.
After selecting $S_t$, the environment first samples outcomes $X_i(t)\sim D_i$ for all $i\in S_t$, and all the random variables $X_i(t)$ (for different $i,t$ pairs) are sampled independently.
Then the environment returns \textit{value-index feedback} $(r_t,i_t)$, where $r_t=\max_{i\in S_t}X_i(t)$ is the maximum outcome, and $i_t=\argmax_{i\in S_t}X_i(t)$ is the index of the arm that achieves this maximum outcome.
Besides, $r_t$ is also the reward of the agent in time step $t$.
Note that ties, shall they occur (we denote this event as $\neg\mathcal{E}_0$), are resolved in an arbitrary manner, although $\mathbb{P}[\neg\mathcal{E}_0]=0$ since $D_i$'s are continuous distributions.
We denote the expected reward of an action $S$ as $r^*(S)$, which is given by:
\begin{align*}
    r^*(S) := \E[\max\{X_i : i \in S\}].
\end{align*}

The objective of the $K$-Max bandits is to select actions $S_t$ properly to maximize the expected cumulative reward $\sum_{t=1}^T r^*(S_t)$ in $T$ rounds.

Let $S^* := \argmax_{S \in \gS} r^*(S)$ denote the optimal action. We evaluate the performance of the agent by the regret metric, which is defined by
\begin{align}\label{eq:def-regret}
    \gR(T) := \E\left[ \sum_{t=1}^T r^*(S^*) - r^*(S_t) \right],
\end{align}
where the expectation is taken over the uncertainty of $\{S_t\}_{t=1}^T$.

\section{Challenges and Resolution in Continuous K-Max Bandits with Value–Index Feedback}
\label{sec:challenges}

To avoid learning over exponentially many super-actions, the key idea of the CMAB framework is to reconstruct per-arm information from the feedback, reducing the statistical burden to per-arm quantities \citep{chen2013combinatorial}.
However, for continuous $K$-Max bandits formulated in \Cref{sec:preliminaries}, the combination of continuous outcome distributions, the \texttt{max} operator, and value-index observability creates obstacles that make previous reduction methods for CMABs fail.

We outline the shortcomings of three existing methods: \textit{(i)} discretization to finite-support settings; \textit{(ii)} parametric Maximum Likelihood Estimation (MLE); and \textit{(iii)} submodularity-driven greedy, and demonstrate how our solution overcomes these limitations.

\subsection{Discretization to finite-support settings}
For {finite-support} $K$-Max bandits with value-index feedback and deterministic tie-breaking (i.e., the winner is deterministic when several arms have the same outcome), expanding each arm into binary sub-arms admits efficient algorithms and $\wt{O}(\sqrt{T})$ regret bounds \citep{wang2023combinatorial}. In the {continuous} case, however, discretization creates {artificial ties} at the top bins, while value-index feedback reveals only a random winner in the same bin; the winner distribution under tie-breaking can vary across continuous outcome distributions (see \Cref{sec:bias-corrected-estimators} for more details). Hence, naive winner-frequency estimators become \emph{selection biased}, which fails the existing analysis based on unbiased estimation of frequencies.

\emph{Our Solution:} We introduce a novel bias-corrected estimation method and develop the concentration analysis incorporating bias-aware error control (\Cref{sec:bias-corrected-estimators}), enabling us to jointly manage estimation variance and discretization-induced bias to achieve efficient learning with biased estimators.

\subsection{Parametric Maximum Likelihood Estimation}
In this way, we often assume $\{f_i(\cdot; \theta_i), F_i(\cdot;\theta_i)\}_{i=1}^N$ to be the probability density function (PDF) and cumulative distribution function (CDF) for each arm, where the vector $\theta_i$ denotes the parameter for arm $i$. For round $t$ with action $S_t$, the feedback $(r_t, i_t)$ has joint density
\begin{equation}
\label{eq:winner-density}
p_\theta(r_t\!=\!r, i_t\!=\!i\mid S_t)=f_i(r;\theta_i)\!\! \prod_{j\in S_t\setminus\{i\}}\!\!F_j(r;\theta_j),
\end{equation}
and the log-likelihood for round $t$ is
\begin{equation}
\label{eq:ri-likelihood}
L_t(\theta)\!=\log f_{i_t}(r_t;\theta_{i_t})+\sum_{j\in S_t\setminus\{i_t\}}\log F_j(r_t;\theta_j).
\end{equation}
To perform online learning analysis for MLE, it must be assumed that the $\{L_t(\theta)\}_t$ is at least embedded into a finite-dimensional statistical manifold.
However, unlike Generalized Linear Bandits (GLB) \citep{liu2024almost,lee2024unified} and Multinomial Logistic (MNL) Bandits \citep{liu2024combinatorial,lee_improved_2024}, the log-likelihood function $L_t$ for $K$-Max bandits \emph{couples both PDFs and CDFs}, which derives inherent hardness for bounding the statistical dimension of $\{L_t(\theta)\}_t$ for general continuous distribution $D_i$.

\textit{Our Solution:} We investigate a structured exponential family where each arm $i \in [N]$ follows an independent exponential distribution. In this case we design a likelihood-based estimator and achieve $\tilde O(\sqrt{T})$ regret (see \Cref{sec:kminexp}).

\subsection{Submodularity-driven Greedy}
Define $f(S)=\E[\max_{i\in S}X_i]$. For any realization $x\in\mathbb{R}^n$, the set function $S\mapsto \max_{i\in S}x_i$ is \emph{monotone submodular}. Since nonnegative linear combinations preserve submodularity, $f$ is monotone submodular.\footnote{This observation also justifies the frequent use of greedy oracles for $K$-cardinality constraints.}
For general monotone submodular objectives, greedy attains the tight $(1-1/e)$ offline ratio under a cardinality constraint \citep{nemhauser1978analysis}. This approximation benchmark is therefore natural in submodular bandit work \citep{yue2011linear,nie2022explore,fourati2024combinatorial}.
However, we consider the regret \Cref{eq:def-regret} measured against the \emph{true} optimum $S^*$.
Then any $\alpha$-approximate oracle with $\alpha<1$ induces a per-round gap of at least $(1-\alpha)\,f(S^*)$, resulting in the \emph{linear} regret.

\textit{Implication for our problem:} The preceding argument demonstrates that a submodularity-driven greedy algorithm is not appropriate for minimizing regret relative to the true optimum, $S^*$, in our scenario. Furthermore, our numerical experiments reveal a linear increase in regret when using the Greedy algorithm (Appendix~\ref{app:experiments}).

\textbf{Preview of our approach.}
Guided by the above arguments, we (i) discretize the outcome space and give a polynomial-time offline oracle for the discrete reduction (\Cref{sec:discretization}); (ii) construct bias-corrected estimators tailored to value-index feedback (\Cref{sec:bias-corrected-estimators}) and provide the polynomial-time algorithm with the first sublinear regret bound (\Cref{alg} in \Cref{sec:algorithm}); and (iii) prove the first $\wt{O}(T^{3/4})$ regret for general continuous $K$-Max Bandits (\Cref{sec:result}). Moreover, for exponential distribution families, we recover the nearly-optimal $\tilde O(\sqrt{T})$ (\Cref{sec:kminexp}).

\section{Algorithm for K-Max Bandits with General Continuous Distribution}
\label{sec:kmax}

We now present our solution framework for continuous $K$-Max bandits, beginning with the fundamental regularity condition that enables discretization-based learning:
\begin{assumption}\label{ass:bi-lipschitz}
Each outcome distribution $D_i$ is supported on $[0,1]$ with a bi-Lipschitz continuous cumulative distribution function (CDF) $F_i$. Specifically, there exists $L \geq 1$ such that for any $i \in [N]$ and $0 \leq v < u \leq 1$:
    $\frac{1}{L}(u - v) \leq F_i(u) - F_i(v) \leq L(u - v).$
\end{assumption}
Many studies on MAB or CMAB consider $[0, 1]$-supported arms \citep{abbasi2011improved,chen2013combinatorial,slivkins2019introduction,lattimore2020bandit}. Assumption~\ref{ass:bi-lipschitz} imposes a simple regularity condition on the continuous outcomes: probability mass over any interval is comparable to the interval length. Similar smoothness conditions are common in continuous bandits \citep{li2017provably,wang2019optimism,liu2023optimistic}, and many compactly supported models, such as mixed uniforms and truncated smooth distributions, satisfy it. We use this assumption to keep the discretized bins controlled and to bound the bias caused by artificial ties.

\subsection{The Discretization of Continuous K-Max Bandits}
\label{sec:discretization}

To derive our algorithm, we first introduce a discretization approach for continuous K-Max Bandits. Discretization is a standard tool in continuum and Lipschitz bandits \citep{kleinberg2004nearly,magureanu2014lipschitz,nika2020contextual}. Under value-index feedback, however, binning continuous outcomes creates artificial ties, and the observed winners no longer give unbiased estimates of the discretized distribution. Our framework uses discretization together with a bias correction term, while also enabling compact storage and computation of continuous distributions, conversion to a set of binary arms, and the use of a polynomial-time offline optimization oracle.

\paragraph{Discretization.}
Since it is complex to estimate the general continuous distributions, a natural idea is to perform \textit{discretization} with granularity $\epsilon$.
Below, we define the \textit{discrete $K$-Max bandits} (called $\bar\gB$) converted from the continuous $K$-Max bandits $\mathcal{B}^*$, where each discrete arm's outcome $\bar X_i$ is discretized from the continuous random variable $X_i$ under granularity $\epsilon$:
    $\bar X_i =  v_j$ if $X_i \in M_j$,
where $M = \ceil{1/\epsilon}$ is the number of discretization bins, $M_j := [(j-1)\epsilon, j\epsilon)$ is the $j$-th bin, and $v_j := (j-1)\epsilon$ is the approximate value of $j$-th bin.
We also let $M_{\le j} = \cup_{j' \le j} M_{j'}$ and $M_{\ge j} = \cup_{j' \ge j} M_{j'}$.
For simplicity, we denote $p^*_{i,j}$ as the probability that $X_i$ falls in $M_j$. That is, for every $i \in [N]$ and $j \in [M]$,
$
    p^*_{i,j} := \Prob[X_i \in M_j] = \Prob[\bar X_i = v_j].
$

Therefore, the discretized problem $\bar \gB$ only depends on the discrete probability set $\vp^* = \{p_{i,j}^* : i \in [N], j \in [M]\}$. Moreover, we set $\bar r(S; \vp^*)$ as the expected reward of an action $S$ in discrete $K$-Max bandits under the probability set $\vp^*$:
\begin{align*}
    \bar r(S ; \vp^*) := \sum_{j\in [M]} v_j \cdot \Prob\left[\max_{i\in S} (\bar X_i) = v_j\right]
\end{align*}
Since $\max_{i\in S} (\bar X_i) = v_j \Rightarrow \max_{i\in S} (X_i)\in M_j$, the key property $\Prob\left[\max_{i\in S} (\bar X_i) = v_j\right] = \Prob\left[\max_{i\in S} (X_i)\in M_j\right]$ gives an upper bound for the discretization error (see formal version in \Cref{lemma:discrete-error-formal}):
\begin{lemma}\label{lemma:discrete-error}
$|r^*(S) - \bar r(S; \vp^*)| \le \epsilon, \quad \forall S \in \gS.$
\end{lemma}
\paragraph{Converting a discrete arm to a set of binary arms.}

Inspired by \citet{wang2023combinatorial}, we decompose a discrete arm $\bar X_i$ into $M$ independent random variables $\{\bar Y_{i,j}\}_{j\in[M]}$, which makes it much easier to understand the estimation method.
Here $\bar Y_{i,j}$ takes value $v_j$ with probability $q^*_{i,j}$ (which is determined by \Cref{eq:qstar-def}) such that $\bar X_i = \max_{j\in[M]}{\bar{Y}_{i,j}}$:
\begin{align}\label{eq:qstar-def}
    q_{i,j}^* := \frac{p_{i,j}^*}{1 - \sum_{j' > j} p_{i,j'}^*}, \ p^*_{i,j} = q^*_{i,j} \cdot \prod_{j' > j} (1 - q^*_{i,j'}).
\end{align}
This conversion allows us to estimate $\vq^* = \{q^*_{i,j}:i\in[N], j\in[M]\}$ instead of $\vp^*$.
For any action $S \in \gS$, define $\bar r_q(S; \vq)$ as the expected maximum reward of $\{\bar Y_{i,j}\}_{i \in S, j \in [M]}$ with probability set $\vq$.
Then we have
\begin{lemma}\label{lemma:r-q-r}
    For any $\vp$ and $\vq$ satisfying \Cref{eq:qstar-def}, we have for any action $S \in \gS$,
    $\bar r_q(S; \vq) = \bar r(S; \vp).$
\end{lemma}
The formal version of this lemma is given in \Cref{lemma:r-q-r-formal}.
Moreover, the function $\bar r_q$ is monotone with respect to  $\vq$, i.e.,
\begin{lemma}[{\citet[Lemma 3.1]{wang2023combinatorial}}]
\label{lemma:monotone}
    For any $\vq'$ and $\vq$ such that $q'_{i,j} \ge q_{i,j}$ holds for any $i \in [N], j \in [M]$, we have
    $\bar r_q(S; \vq') \ge \bar r_q(S;\vq)$ for any $S \in \gS$.
\end{lemma}
\paragraph{Polynomial approximation offline oracle.}
For any discrete $K$-Max bandits with probability set $\vp$, we can apply the \textit{PTAS} algorithm \citep{chen2016combinatorial}
as a polynomial time offline ($1-\epsilon$)-approximation optimization oracle for any given $\epsilon \in (0, 1)$.
Moreover, for any probability set $\vq$, we can convert it to $\vp$ by \Cref{eq:qstar-def}, input this $\vp$ to the PTAS oracle and get the approximation solution $\operatorname{PTAS}(\vp)$ satisfying
\begin{equation}\label{eq:ptas}
\begin{aligned}
    &\quad \bar r_q\left(\operatorname{PTAS}(\vp); \vq\right) = \bar r\left(\operatorname{PTAS}(\vp); \vp\right) \\
    &\ge (1-\epsilon) \cdot \max_{S \in \gS} \bar r(S; \vp) = (1-\epsilon) \cdot \max_{S \in \gS} \bar r_q(S; \vq),
\end{aligned}
\end{equation}
which provides a way to control the relative error of the PTAS oracle.

\subsection{\texttt{DCK-UCB}: The Efficient Continuous K-Max Bandits Algorithm} \label{sec:algorithm}

Building on the methodology in the previous subsection,
we adapt the framework in \citet{wang2023combinatorial}, and
present \texttt{DCK-UCB} (Discretized Continuous $K$-Max with Upper Confidence Bounds, with pseudo-code in \Cref{alg}), the first efficient algorithm addressing $K$-Max bandits with general continuous outcome distributions.
At a high level, we first discretize the continuous $K$-Max bandits into discrete $K$-Max bandits. Then we convert every discrete arm to a set of binary arms and estimate the corresponding $\vq^*$ parameters by $\bar{\vq}$. Finally, we convert $\bar{\vq}$ back to $\bar{\vp}$, input $\bar{\vp}$ to the $\operatorname{PTAS}$ oracle, and get the action we want to select.

\begin{algorithm}[t]
\caption{\texttt{DCK-UCB}: Discretized Continuous $K$-Max Bandits with Upper Confidence Bonus}
\label{alg}
\begin{algorithmic}[1]
\REQUIRE Discretization granularity $\epsilon$ and the offline $(1-\epsilon)$-approximated optimization oracle $\operatorname{PTAS}$ for discrete $K$-Max bandits \citep{chen2016combinatorial}.
\STATE Initialize $M \leftarrow \ceil{1/\epsilon}$, $\hat q_{i, 1}^1 \leftarrow 1$ for every $i \in [N]$, and $\hat q_{i,j}^1 \leftarrow 0$ for every $i\in [N], j > 1$.
\FOR{$t=1,2,\ldots,T$}
\STATE For every $i \in [N], j \in [M]$, with confidence bonus $\beta_{i,j}^t$ defined in \Cref{eq:def-beta}, we set
\begin{align}\label{eq:def-barq}
    \bar q_{i, j}^t \leftarrow \min\left\{ \hat q_{i,j}^t + \beta_{i,j}^t + (K-1)\frac{L^4}{j^2}, 1\right\}.
\end{align}
\STATE Convert $\bar \vq^t$ to $\bar \vp^t$ by \Cref{eq:qstar-def}.
\STATE \label{algline:oracle} Choose action $S_t \leftarrow \operatorname{PTAS}(\bar \vp^t)$.
\STATE Observe $(r_t,i_t)$ by executing action $S_t$. Denote $j_t$ as the bin index of $r_t$, i.e., $r_t \in M_{j_t}$.
\STATE For any $i, j \in [N] \times [M]$,
\vspace{-.2cm}
\begin{align*}
    C_t(i, j) &\leftarrow
        C_{t-1}(i, j) + \1\left[ i = i_t \And j = j_t \right], \\
    SC_t(i, j) &\leftarrow
        SC_{t-1}(i, j) + \1\left[ i \in S_t \And j \ge j_t \right].
\end{align*}
\STATE Calculate estimator $\hat q_{i,j}^{t+1} \leftarrow \frac{C_t(i,j)}{SC_t(i,j)}$, for every $i \in [N]$ and $j\in [M]$.
\ENDFOR
\end{algorithmic}
\end{algorithm}

In Line 3 of \Cref{alg}, we calculate the optimistic estimator $\bar q_{i,j}^t$ which upper bounds $q^*_{i,j}$ with high probability.
This is done by adding two upper confidence bonus terms $\beta_{i,j}^t$ and $(K-1)L^4/j^2$.
Analysis shows that $\bar q_{i,j}^t \ge q^*_{i,j}$ with high probability (Lemma \ref{lemma:concentration}). The detailed discussion on this estimator will be given in the next subsection.
In Lines 4-5, the agent converts this $\bar \vq$ to $\bar \vp$, and then runs the offline $(1-\epsilon)$-approximation optimization oracle $\operatorname{PTAS}$
to get action $S_t$ for execution.
In Line 6, the agent gets the value-index return $(r_t, i_t)$, and discretizes the value $r_t$ to the bin index $j_t$, i.e., $r_t \in M_{j_t}$.
In Lines 7-8, the agent estimates $\vq^*$ by two counters: $C_t(i,j)$ counts the times when $(i,j)$ exactly equals the feedback $(i_t,j_t)$, and $SC_t(i,j)$ counts the number of steps $\tau \le t$ satisfying $i \in S_\tau$ and $j_\tau \le j$. Moreover, \texttt{DCK-UCB} achieves polynomial complexity, with $\mathrm{poly}(N,K,1/\epsilon)$ per round time and $\gO(N/\epsilon)$ space.

\subsection{Bias-Corrected Estimators} \label{sec:bias-corrected-estimators}
The key challenge in the algorithm design and theoretical analysis is that $\hat q_{i,j}^t$ is not an unbiased estimator for $q_{i,j}^*$. This means that except for the confidence radius due to the randomness of the environment, we still need another bonus term to bound the bias to guarantee that $\bar q_{i,j}^t$ is a UCB for $q_{i,j}^*$.
Specifically, note that
\begin{align*}
    q^*_{i,j}
    = \frac{p^*_{i,j}}{1 - \sum_{j' > j} p^*_{i,j'}}
    = \frac{p^*_{i,j}}{\sum_{j'=1}^j p^*_{i,j'}}
    = \frac{\Prob[X_i \in M_j]}{\Prob[X_i \in M_{\le j}]}.
\end{align*}
If we have an assumption that when $i\in S_\tau$ and $j_\tau = j$, $X_i(\tau) \in M_j$ implies $i = i_\tau$, then we can guarantee that $\hat q_{i,j}^{t} = C_t(i,j) / SC_{t}(i,j)$ is an unbiased estimator for $q^*_{i,j}$. This is because in this case,
\begin{align*}
    \frac{C_t(i,j)}{ SC_{t}(i,j) }&= \frac{\# \text{ of } i_\tau = i \And j_\tau = j}{\# \text{ of } i\in S_\tau \And j_\tau \le j} \\
    &= \frac{\# \text{ of } X_i(\tau) \in M_j \And i\in S_\tau \And j_\tau \le j}{\# \text{ of } i\in S_\tau \And j_\tau \le j},
\end{align*}
which is the fraction of $X_i(\tau) \in M_j$ conditioned on $i\in S_\tau, j_\tau \le j$,
and is an unbiased estimator for
\begin{align*}
    &  \Prob[X_i(\tau) \in M_j \mid i\in S_\tau, j_\tau \le j] \\
    =  &\frac{\Prob[X_i(\tau) \in M_j, i\in S_\tau, j_\tau \le j]}{\Prob[i\in S_\tau, j_\tau \le j]}\\
    =&\frac{\Prob[X_i(\tau)  \in M_j]\cdot \Prob[X_k(\tau)  \in M_{\le j}, \forall k \in S_{\tau}, k \neq i]}{\Prob[X_i(\tau)  \in M_{\le j}] \cdot \Prob[X_k(\tau)  \in M_{\le j}, \forall k \in S_{\tau}, k \neq i]}\\
    =&\frac{\Prob[X_i(\tau) \in M_j]}{\Prob[X_i(\tau) \in M_{\le j}]}
\end{align*}

However, we know that in the discrete K-Max bandits converted from the continuous K-Max bandits, there is no such assumption (different from \citet{wang2023combinatorial} who requires deterministic tie-breaking rule). When multiple arms have $X_i(\tau) \in M_j$, the observed winning arm $i_\tau = \arg\max X_i(\tau)$ is not a fixed one, and even we do not know the distribution of the winner.
Because of this, we cannot guarantee that conditional on $i\in S_\tau, j_\tau \le j$, we increase the counter $C_t(i, j)$ for every time step $\tau$ such that $X_i(\tau) \in M_j$. Some steps where $X_i(\tau) \in M_j$ but $X_i(\tau)$ is not the winner are missed.
This nondeterministic tie-breaking effect, arising from the continuous-to-discrete transformation, induces systematic negative bias in conventional estimators $\{\hat q_{i,j}^t\}$.
Therefore, to guarantee that $\{\bar q_{i,j}^t\}$ is an upper confidence bound of $\{ q_{i,j}^*\}$, we need another bonus term.

Specifically, this required bonus term arises because, conditional on $i\in S_\tau, j_\tau \le j$, there are some time steps where $X_i(\tau) \in M_j$ but arm $i$ is not the winner and thus we miss these steps in counter $C_t(i, j)$.
When this happens, there must be at least one other arm $i'\ne i, i'\in S_{\tau}$ such that $X_{i'}(\tau) \in M_j$.
This probability can be upper bounded by
\begin{align*}
    &\sum_{i'\ne i, i'\in S_{\tau}} \Prob[X_i(\tau) \in M_j, X_{i'}(\tau) \in M_j \mid i\in S_\tau, j_\tau \le j]\\
    &=\sum_{i'\ne i, i'\in S_{\tau}} \frac{p^*_{i,j}p^*_{i',j} }{ \sum_{j'\le j}p^*_{i,j'} \sum_{j'\le j}p^*_{i',j'}}\\
    &\le (K-1)\frac{(L\epsilon)^2}{(j\epsilon/L)^2}
    = (K-1)\cdot(L^4/j^2),
\end{align*}
where the last inequality follows from $p^*_{i,j}\le L\epsilon$ and $\sum_{j'\le j}p^*_{i,j'}\ge j\epsilon/L$ under \Cref{ass:bi-lipschitz}.
Thus \Cref{ass:bi-lipschitz} converts the artificial-tie probability into the explicit bias-correction term for every $q_{i,j}^*$.

\begin{lemma}\label{lemma:concentration}
Under \Cref{ass:bi-lipschitz}, let the confidence radius be defined as
\begin{align}\label{eq:def-beta}
    \beta_{i,j}^t := \sqrt{8\frac{\log(NMT)}{SC_{t-1}(i,j)}}.
\end{align}
Then with probability at least $1 - T^{-2}$,
$$
    \left|\hat q_{i,j}^t - q^*_{i,j}\right| \le {\beta_{i,j}^t} + (K-1)\cdot(L^4/j^2),
$$
holds for every $t \in [T]$, $i \in [N]$ and $j \in [M]$.
\end{lemma}

\subsection{Regret Analysis and Discussion}
\label{sec:result}
We establish the first efficient algorithm \texttt{DCK-UCB} (\Cref{alg}) which enjoys sublinear regret guarantees in the continuous $K$-Max bandits problem with value-index feedback.
\begin{theorem}
\label{thm:main}
Under \Cref{ass:bi-lipschitz},
given the exploration bonus term $\beta_{i,j}^t$ in \Cref{eq:def-beta}, discretization granularity $\epsilon = \gO(L^{-2}K^{-3/4}N^{1/4}T^{-1/4})$ and PTAS approximation factor $\alpha = 1 - \epsilon$, \Cref{alg} enjoys the regret guarantee
    $\gR(T) \le \wt{\gO}(L^{2}N^{1/4}K^{5/4}T^{3/4}).$
\end{theorem}

\Cref{thm:main} establishes a regret bound sublinear-in-$T$ for our DCK-UCB algorithm. To the best of our knowledge, this is the first sublinear regret guarantee for the continuous K-Max bandit problem under the challenging value-index feedback model.

The key component of achieving \Cref{thm:main} is the novel bias-corrected estimators and the concentration analysis with bias-aware error control, as detailed in \Cref{sec:bias-corrected-estimators}.
The analysis divides the cumulative regret into two parts: the estimation error which is upper bounded by the summation of the exploration bonus term $\beta_{i,j}^t$, and the bias-correction error which is upper bounded by the summation of the bias-correction term $(K-1)\cdot(L^4/j^2)$.
The first one is upper bounded by $\wt{\gO}\left(\sqrt{NKT}/\epsilon\right)$, following a regular CMAB-type analysis \citep{wang2017improving} where there are $NM \approx N/\epsilon$ bins in total and the player needs to pull $KM \approx K/\epsilon$ of them in each time step.
The second term, on the other hand, needs to take summation over all the pulled $KM \approx K/\epsilon$ bins in every time step, while the bias-correction term for the $j$-th bin may influence the regret by at most $j\epsilon$.
Thus, it is upper bounded by $\wt{\gO}\left( K^2L^4T\epsilon\right)$.
By choosing the discretization granularity $\epsilon$ for balancing these two terms, we achieve the regret guarantee in \Cref{thm:main}, which is the first theoretical result sublinear in $T$ for continuous $K$-Max bandits.
Please refer to \Cref{thm:main-formal} for the formal statement and complete proof.

\paragraph{Comparison to Prior Works.} Our $\wt{\gO}(T^{3/4})$ regret bound advances the state-of-the-art in several ways. While \citet{wang2023combinatorial} achieves $\gO(S\sqrt{T})$ regret for discrete $K$-Max bandits with finite support size $S$, their approach does not work efficiently in the continuous case due to the infinite support size.
Although using our discretization method can address the issue of infinite support size, their analysis still fails due to the non-deterministic tie-breaking problem and the biased estimators, which is one of the key contributions of our work.

Recent submodular bandits work \citep{pasteris2023sum,fourati2024combinatorial, tajdini2024nearly} reaches $\wt{\gO}(T^{2/3})$ regret but has key limitations: their regret baseline is an $(1-1/e)$-approximation, which leads to a linear regret in our definition. Moreover, they require flexibility in subset size selection. Our framework overcomes these challenges through bias-corrected estimators with PTAS integration, effectively handling continuous $K$-Max bandits.
\paragraph{Discussion on the regret order in \Cref{thm:main}.}
The regret bounds established in \Cref{thm:main} result from strategically selecting the discretization granularity $\epsilon$ to achieve an optimal balance between the estimation error and bias-correction error.
The $\sqrt{NKT}$ factor appearing in the upper bound of exploration bonus term $\wt{\gO}(\sqrt{NKT/\epsilon^2})$ aligns with established $\Omega(\sqrt{NKT})$ lower bounds in combinatorial multi-armed bandits \cite{kveton2015tight}. Regarding the upper bound of bias-correction error, our analysis in \Cref{sec:bias-corrected-estimators} shows that the estimation bias of $q^*_{i,j}$ is approximately the probability of another arm's outcome falling into the same discretized bin $j$, which leads to a bound of $\gO(KL^4/j^2)$.
Moreover, a positive bias in $q^*_{i,j}$ may cause the reward to be overestimated by at least $\epsilon$.
When aggregated across all $K$ arms in the selected action $S_t$ and all $t \le T$, this yields the $\wt\gO(K^2L^4 T\epsilon)$ term, which seems inevitable.

In summary, we argue that these dependencies on $N,K$ and $T$ in both error components accurately capture the problem's fundamental complexity.
Nevertheless, improving the relationship between discretization granularity $\epsilon$ and estimation error, e.g., reducing its dependency on $\epsilon$ from the current $\wt{\mathcal{O}}(\sqrt{NKT/\epsilon^2})$ to $\wt{\mathcal{O}}(\sqrt{NKT/\epsilon})$, could potentially enable a more favorable balance between these terms, leading to tighter bounds on $N$, $K$, and $T$.
This is a promising yet challenging direction towards tighter regret bounds.
Previous studies in combinatorial MABs \citep{liu2023contextual,liu2024combinatorial} suggest a variance-adaptive analysis for the exploration bonus terms. However, achieving this faces substantial technical challenges, primarily because the observations are not i.i.d. due to the non-deterministic tie-breaking issue, which introduces new difficulties in deriving concentration inequalities for the variance terms of these biased observations.

\section{Better Performance under the Exponential Distributions}
\label{sec:kminexp}

In this section, we demonstrate how specific distributional structure enables the improvement of the regret guarantee from $\wt{\gO}(T^{\frac{3}{4}})$ to $\wt{\gO}(\sqrt{T})$. We investigate the special case of $K$-Min exponential bandits, which can be viewed as a variant of continuous $K$-Max bandits where we seek to minimize losses rather than maximize rewards. For space limitations, we put the detailed discussion in Appendix~\ref{app:kminexp}.

In the $K$-Min exponential bandits problem, each arm $i$ generates a loss $X_i$ from an exponential distribution $X_i \sim D_i = \Exp(\mu_i)$ where $\mu_i > 0$. The agent selects a subset $S_t \in \mathcal{S} = \{S \subseteq [N] : |S| = K\}$ and observes only the minimum loss $\ell_t = \min_{i \in S_t} X_i(t)$ (full bandit feedback without the winner's index).\footnote{This can be transformed into a $K$-Max problem by setting $Z_i(t) = -X_i(t)$ and viewing $Z_i(t)$ as rewards.}
Similar to previous works on general linear bandits \citep{liu2024almost, lee2024unified, liu2024combinatorial}, we assume a linear structure where $\mu_i = \langle \phi(i), \theta^* \rangle$ for some unknown parameter $\theta^* \in \Theta \subset \mathbb{R}^d$ and a known bounded feature mapping $\phi: [N] \rightarrow \mathbb{R}^d$.

The critical insight that enables improved regret bounds is that the minimum of exponential random variables follows an exponential distribution with a parameter equal to the sum of the individual parameters, i.e.,
\begin{align*}
    \min_{i \in S} X_i \sim \Exp\left(\sum_{i \in S} \mu_i\right) = \Exp\left(\sum_{i \in S} \langle \phi(i), \theta^*\rangle \right)
\end{align*}
This property allows us to directly estimate $\theta^*$ using maximum likelihood estimation without requiring discretization or value-index feedback.
Specifically, let $\psi(S)  := \sum_{i \in S} \phi(i)$, $\forall S \in \gS$. Then with chosen action $S_t$ and parameter $\theta$, the observed loss should follow the exponential distribution $\Exp\left(\sum_{i \in S} \phi(i)^T \theta \right) = \Exp\left(\psi(S)^T\theta\right)$, whose probability density function is $f(x) = (\psi(S)^T\theta) \cdot e^{\left(-(\psi(S)^T\theta)  x\right)}$.
Because of this, the log-likelihood function is
\begin{equation}\label{eq:kmin-loglikelihood}
\begin{aligned}
    L_t(\ell_t; S_t, \theta) :&=  \log \left( \psi(S_t)^\top \theta e^{\left( -\psi(S_t)^\top \theta \ell_t\right)} \right).
\end{aligned}
\end{equation}
Note that this is a $K$-Min extension of \Cref{eq:ri-likelihood} when the winner's index is not given. Specifically, let $f_\mu(x), F_\mu(x)$ be the PDF and CDF for exponential distribution with rate $\mu$, \Cref{eq:ri-likelihood} gives that
	    $L_t(\theta) = \log (\sum_{i \in S_t} f_{\phi(i)^\top\theta}(\ell_t) \prod_{j \in S_t \setminus \{i\}} (1 - F_{\phi(j)^\top\theta}(\ell_t)))
	    = \log (\sum_{i \in S_t} \phi(i)^\top\!\theta  e^{-\phi(i)^\top\!\theta\ell_t} \!\prod_{j \in S_t \setminus \{i\}} e^{-\phi(j)^\top\theta\ell_t}),$
which is exactly \Cref{eq:kmin-loglikelihood}.
This allows us to apply the MLE method.
Denote $\gL_t(\theta)$ as the summation of $L_t$ and a regularization term with factor $\lambda$:
\begin{align}\label{eq:loglikelihood-def}
    \gL_t(\theta; \lambda) := \sum_{i < t} - L_i(\ell_i; S_i, \theta) + \frac{\lambda}{2}\|\theta\|^2,
\end{align}
Then we are ready to present the algorithm \texttt{MLE-Exp} (\Cref{alg:k-min}), which constructs confidence sets around the MLE estimate and selects actions that minimize the expected loss.
\begin{algorithm}[htbp]
\caption{\texttt{MLE-Exp}: MLE for $K$-Min Exponential Bandits}
\label{alg:k-min}
\begin{algorithmic}[1]
\REQUIRE Regularization factors $\{\lambda_t\}_{t\in [T]}$, confidence radius $\{\gamma_t\}_{t\in [T]}$, and probability constant $\delta$.
\FOR{$t = 1, \ldots, T$}
    \STATE Compute MLE $\hat\theta_t$ by
    \[\hat{\theta}_t \leftarrow \argmin_{\theta \in \R^d} \mathcal{L}_t(\theta; \lambda_t),\]
    where $\gL_t(\theta;\lambda)$ is given in \Cref{eq:loglikelihood-def}.
	    \STATE Construct the confidence set $C_t(\hat\theta_t; \delta, \lambda_t)$ according to \Cref{eq:def-C-set}
	    \STATE Select action with minimum expected loss:
	    $$(S_t, \wt{\theta}_t) \leftarrow \argmax_{S \in \gS, \theta \in C_t(\hat{\theta}_t; \delta, \lambda_t)} \langle \psi(S), \theta \rangle.$$
	    \STATE Play action $S_t$ and observe the loss $\ell_t$.
\ENDFOR
\end{algorithmic}
\end{algorithm}

In Line 2 of \Cref{alg:k-min}, we estimate the MLE $\hat\theta_t$ by minimizing the summation of the log-likelihood function and the regularization term $\gL_t(\theta, \lambda_t)$.
Given $\lambda_t$ a priori, we will write $\gL_t(\theta)$ instead of $\gL_t(\theta, \lambda_t)$ for simplicity.
Inspired by \citet{liu2024almost, lee2024unified, liu2024combinatorial}, in Line 3, we construct a confidence set $C_t(\hat\theta_t;\delta)$ (defined in \Cref{eq:def-C-set}), centered at the MLE $\hat\theta_t$ with confidence radius $\gamma_t(\delta)$, based on the gradient term $g_t(\theta) := -\nabla_\theta \gL_t(\theta) + \sum_{i < t} \ell_i \psi(S_i)$ and Hessian matrix $H_t(\theta) := \nabla^2_\theta \gL_t(\theta)$:
\begin{align}\label{eq:def-C-set}
    C_t(\hat\theta_t; \delta) :=
    &\left\{ \theta \in \Theta : \left\|g_t(\theta) - g_t(\hat\theta_t)\right\|_{H_t^{-1}(\theta)} \le \gamma_t(\delta)\right\},
\end{align}
\Cref{lemma:mle-concentration} shows that with probability at least $1-\delta$, we have $\theta^* \in C_t(\hat\theta_t; \delta)$.
Then in Line 4, we apply a double oracle to look for the action $S$ whose expected loss under a parameter $\theta$ in the confidence set ($1/\langle \psi(S), \theta \rangle$) is minimized.
Finally, we select this greedy action in Line 5 and use the observation to update the next time step's MLE and confidence set.

The following theorem shows that the \texttt{MLE-Exp} algorithm achieves a $\wt{O}(\sqrt{T})$ regret upper bound:
\begin{theorem}
\label{thm:kminexp}{\texttt{MLE-Exp}} achieves regret upper bound $\mathcal{R}(T) \leq \widetilde{\gO}\left(\sqrt{d^3 T}\right).$
\end{theorem}
This theorem represents a significant improvement over the $\widetilde{\gO}(T^{3/4})$ bound for general continuous $K$-Max bandits.
Combined with the $\Omega(\sqrt{T})$ lower bound proven in Appendix~\ref{app:lowerbound}, we show that \texttt{MLE-Exp} achieves the nearly minimax optimal regret with respect to $T$, even without winner's index feedback.
Note that the regret bound in \Cref{thm:kminexp} is derived under linear parametric assumptions, which explains why the bound depends on the feature dimension $d$ rather than explicitly on $N$ or $K$. This dimension $d$ inherently encapsulates information about the problem structure, including the number of arms and selection size, which aligns with established results in other combinatorial linear bandit settings \citep{liu2023contextual,liu2024combinatorial}. The detailed proof of Theorem \ref{thm:kminexp} is provided in Appendix~\ref{Appendix:k-min}.

Exponential $K$-Min bandits form a special case of continuous $K$-Max bandits. In \Cref{sec:kmax}, the general solution for continuous $K$-Max bandits reaches a regret bound of $\widetilde{\gO}(T^{3/4})$ by handling discretization and bias correction under value-index feedback. In contrast, \Cref{thm:kminexp} uses the exponential structure to obtain a near-optimal regret bound even under weaker full-bandit feedback. This suggests that additional distributional structure can lead to sharper algorithms and guarantees.

\section{Conclusion and Future Work}

We studied Continuous $K$-Max Bandits with value–index feedback, a challenging setting due to continuous outcomes, discretization error, and biased observations. We proposed \texttt{DCK-UCB}, the first efficient algorithm with a sublinear $\widetilde{\mathcal{O}}(T^{3/4})$ regret (\Cref{thm:main}), and validated it with numerical experiments (Appendix~\ref{app:experiments}). For the special case of exponential distributions, we introduced \texttt{MLE-Exp}, achieving a near-optimal $\widetilde{\mathcal{O}}(\sqrt{T})$ regret (\Cref{thm:kminexp}) that matches the minimax lower bound (proven in Appendix~\ref{app:lowerbound}).

\textbf{Future Directions:}
A key direction is improving the $\wt{\gO}(T^{3/4})$ bound for general continuous distributions. Variance-aware algorithms, which exploit second-order statistics as in CMABs \citep{liu2023contextual,liu2024combinatorial}, may reduce regret to $\wt{\cO}(T^{2/3})$.
However, such approaches face inherent challenges due to the biased estimations induced by nondeterministic tie-breaking, which create new concentration challenges for variance terms of biased observations. Overcoming these limitations may require developing new bias-corrected concentrations for variance estimators or alternative feedback models tailored to continuous outcomes.
Further avenues include establishing lower bounds for $K$-Max bandits and relaxing the bi-Lipschitz assumption.

\bibliography{ref}

\newpage

\onecolumn

\begingroup
\makeatletter
\def\@thanks{}
\makeatother
\setcounter{footnote}{0}
\title{On the Sublinear Regret of Continuous K-Max Bandits\\(Supplementary Material)}
\maketitle
\endgroup

\appendix

\startcontents[section]
\printcontents[section]{l}{1}{\setcounter{tocdepth}{2}}
\newpage

\section{Omitted Proofs in Section~\ref{sec:kmax}}

In this section, we present the omitted proofs in \Cref{sec:kmax}, which include the full proof of \Cref{thm:main}.

\subsection{Discretization Error}\label{app:discrete-error}
First we show that the discretization from the original continuous problem $\gB^*$ to $\bar\gB$ with discretization width $\epsilon$ incurs a controllable error in expected loss, which is shown in \Cref{lemma:discrete-error} and formalized by the following lemma.

\begin{lemma}[Formal version of \Cref{lemma:discrete-error}]\label{lemma:discrete-error-formal}
For any $S \in \gS$, we have
\begin{align}\label{eq:discrete-error}
    \bar r(S; \vp^*) \le r^*(S) \le \bar r(S; \vp^*) + \epsilon.
\end{align}
\begin{proof}
Notice that we have
\begin{align*}
    \Prob\left[\max_{i\in S} (\bar X_i) = v_j\right] &= \sum_{I\subset S} \prod_{i\in I}\Prob[\bar X_i = v_j] \cdot \prod_{k \in S, k \notin I}\Prob[\bar X_k < v_j] \\
    &= \sum_{I\subset S} \prod_{i\in I} \Prob[X_i \in M_j] \cdot \prod_{k \in S, k \notin I}\Prob[X_k \in M_{\le j-1}] \\
    &= \Prob\left[\max_{i \in S} (X_i) \in M_j\right].
\end{align*}
Therefore, by definition of $r^*(S)$, we have
\begin{align*}
    r^*(S) &= \sum_{j\in [M]} \int_{r \in M_j} r\cdot \rd\Prob_{\max_{i\in S}(X_i)}(r) \\
    &\ge \sum_{j\in [M]} (j-1)\epsilon \int_{r \in M_j} \rd\Prob_{\max_{i\in S}(X_i)}(r) \\
    &= \sum_{j\in [M]} (j-1)\epsilon \cdot \Prob\left[\max_{i \in S} (X_i) \in M_j\right] \\
    &= \sum_{j\in [M]} (j-1)\epsilon \cdot \Prob\left[\max_{i\in S} (\bar X_i) = (j-1)\epsilon\right] \\
    &= \bar r(S; \vp^*),
\end{align*}
where the inequality is given by the definition of $M_j$. This proves the left-hand side of \Cref{eq:discrete-error}. For the other side, we can similarly establish
\begin{align*}
    r^*(S) &= \sum_{j\in [M]} \int_{r \in M_j} r\cdot \rd\Prob_{\max_{i\in S}(X_i)}(r) \\
    &\le \sum_{j\in [M]} j\epsilon \int_{r \in M_j} \rd\Prob_{\max_{i\in S}(X_i)}(r) \\
    &= \sum_{j\in [M]} (j - 1)\epsilon \cdot \Prob\left[\max_{i \in S} (X_i) \in M_j\right] + \epsilon \cdot  \sum_{j\in [M]}  \Prob\left[\max_{i \in S} (X_i) \in M_j\right]\\
    &= \bar r(S; \vp^*) + \epsilon.
\end{align*}
\end{proof}
\end{lemma}

\subsection{Converting to Binary Arms}
As detailed in \Cref{sec:discretization}, we set
\begin{align*}
    q_{i,j}^* := \frac{p_{i,j}^*}{1 - \sum_{j' > j} p_{i,j'}^*}, \ p^*_{i,j} = q^*_{i,j} \cdot \prod_{j' > j} (1 - q^*_{i,j'}),
\end{align*}
which implies
\begin{align*}
    q^*_{i,j} = \frac{p^*_{i,j}}{1 - \sum_{j' > j} p^*_{i,j'}} = \frac{p^*_{i,j}}{\sum_{j'=1}^j p^*_{i,j'}} =
    \frac{\Prob[X_i \in M_j]}{\Prob[X_i \in M_{\le j}]}.
\end{align*}
For any given probability set $\vq = \{q_{i,j}: i \in [N], j \in [M]\}$, we can apply \Cref{eq:qstar-def} to get the corresponding $\vp$ defined as
\begin{align*}
    p_{i,j} = q_{i,j} \cdot \prod_{j' > j} (1 - q_{i,j'}).
\end{align*}
Assume $\{Y_{i,j}^\vq\}_{i\in[N], j \in [M]}$ is the set of independent binary random variables such that $Y_{i,j}^\vq$ takes value $v_j = (j-1)\epsilon$ with probability $q_{i,j}$ and takes value $0$ otherwise. Also let $\{X_i^\vp\}_{i\in[N]}$ be the set of independent discrete random variables such that $X_i^\vp$ takes value $v_j$ with probability $p_{i,j}$ for every $j \in [M]$. Therefore, by simple calculation, $\max_{j\in[M]}\{Y_{i,j}^\vq\}$ has the same distribution as $X_i^\vp$.

$\bar r_q(S; \vq)$ is defined as the expected maximum reward of $\{Y_{i,j}^\vq\}_{i\in S,j\in[M]}$. Then we can write
\begin{align}\label{eq:def-rq}
   \bar r_q(S; \vq) = \sum_{j \in [M]} v_j \cdot \left( Q_j(S; \vq) - Q_{j-1}(S;\vq)\right),
\end{align}
where we denote for simplicity
\begin{align}\label{eq:def-Qj}
    Q_j(S; \vq):= \prod_{k \in S, j' > j} (1 - q_{k,j'}).
\end{align}
$Q_j(S; \vq)$ is actually the probability of the event that every arm in $\{\bar Y_{k,j'}\}_{k \in S, j' > j}$ does not sample a non-zero value.

Equipped with the above statement, we can establish the following lemma:

\begin{lemma}[Formal version of \Cref{lemma:r-q-r}]\label{lemma:r-q-r-formal}
For any $\vp = \{p_{i,j}\}_{i\in[N],j\in[M]}$ and $\vq=\{q_{i,j}\}_{i\in[N],j\in[M]}$ satisfying
\[
    q_{i,j} := \frac{p_{i,j}}{1 - \sum_{j' > j} p_{i,j'}}, \ p_{i,j} = q_{i,j} \cdot \prod_{j' > j} (1 - q_{i,j'}),
\]
we have for any $S \in \gS$, $$\bar r_q(S; \vq) = \bar r(S; \vp).$$
\begin{proof}
    Notice that by definition, we have
    \begin{align*}
        \bar r(S; \vp) = \E\left[\max_{i\in S} X_i^\vp\right],
    \end{align*}
    and
    \begin{align*}
        \bar r_q(S; \vq) = \E\left[\max_{i \in S} \max_{j \in [M]} Y_{i,j}^\vq\right].
    \end{align*}
    Notice that $\max_{j\in[M]}\{Y_{i,j}^\vq\}$ has the same distribution as $X_i^\vp$, we have
    \begin{align*}
        \bar r_q(S; \vq) &= \E\left[\max_{i \in S} \max_{j \in [M]} Y_{i,j}^\vq\right] \\
        &= \E\left[\max_{i \in S} X_i^\vp\right] = \bar r(S; \vp).
    \end{align*}
\end{proof}
\end{lemma}

\subsection{Biased Concentration}
We aim to use $\hat q_{i,j}^t$ to estimate $q_{i,j}^*$. However, this is a biased estimation. In this section, we carefully control the gap between the biased estimator $\hat q_{i,j}^t$ and the true probability $q^*_{i,j}$.

We set $c_t(i, j) := \mathbbm{1}[(i_t, j_t) = (i, j)]$, which is $\gF_t$-measurable. Then \Cref{alg} counts the summation of $c_t(i,j)$ as $C_t(i,j)$:
\begin{align*}
    C_t(i,j) = \sum_{\tau = 1}^t c_\tau(i,j),
\end{align*}
which is $\gF_{t-1}$-measurable.

For a given action $S_t$ in round $t$, the environment samples a set of outcomes $\{X_i(t) \sim D_i : i \in S_t\}$. The value-index feedback is $r_t = \max_{i \in S_t} X_i(t)$, $i_t = \argmax_{i \in S_t} X_i(t)$. \Cref{alg} considers $j_t$ such that $r_t \in M_{j_t}$. We denote $I_t = \argmax_{i \in S_t} \bar X_i(t)$, where $\bar X_i(t)$ is the discretized version of $X_i(t)$. Notice that under event $\gE_0$, $\argmax_{i \in S} X_i(t)$ is unique. But $I_t$ might be a set with multiple indices. We emphasize that $S_t$ is $\gF_{t-1}$-measurable and $(i_t, r_t, j_t, I_t)$ are $\gF_t$-measurable.

Then we can provide the following lemma.
\begin{lemma}[Formal version of \Cref{lemma:concentration}]\label{lemma:concentration-formal}
Under event $\gE_0$, we have for every $t \in [T]$ and $(i, j) \in [N] \times [M]$,
\begin{align*}
    \left|\hat q_{i,j}^t - q^*_{i,j}\right| \le \sqrt{8\frac{\log(NMT)}{SC_t(i,j)}} + (K-1)\cdot(L^4/j^2),
\end{align*}
with probability at least $1 - T^{-2}$, where we denote this good event as $\gE_1$.

\begin{proof}
Denote $q_{i,j}(S_t) := \mathbbm{1}[i \in S_t] \cdot \Prob[(i_t, j_t) = (i, j) \mid j_t \le j, S_t]$, and $q_{i,j}^*(S_t) := \mathbbm{1}[i \in S_t] \cdot \Prob[I_t \ni i, j_t = j \mid j_t \le j, S_t]$. Therefore, for given $i \in [N], j \in [M]$, we have
\begin{align*}
    \E[\1[i \in S_t] \cdot c_t(i,j)\cdot \mathbbm{1}[j_t \le j] \mid  S_t] = q_{i,j}(S_t) \cdot \Prob[j_t \le j \mid S_t]
\end{align*}

By summation over time step $1, 2, \cdots, t$, we have
\begin{align*}
    \sum_{\tau=1}^t \E[\1[i \in S_\tau]\cdot c_\tau(i,j)\cdot \mathbbm{1}[j_\tau \le j] \mid  S_\tau] &= \sum_{\tau=1}^t q_{i,j}(S_\tau)\cdot \Prob[j_\tau \le j \mid S_\tau] \\
    &= \sum_{\tau=1}^t \E\left[q_{i,j}(S_\tau)\cdot \mathbbm{1}[j_\tau \le j] \mid S_\tau\right],
\end{align*}
which implies that
\begin{align*}
     \E\left[\sum_{\tau \le t, i \in S_\tau, j_\tau \le j}c_\tau(i,j) \middle| S_1, S_2, \cdots, S_t \right]  = \E\left[\sum_{\tau \le t, j_\tau \le j}
    q_{i,j}(S_\tau)\middle| S_1, \cdots, S_t\right]
\end{align*}
Notice that $S_t$ is $\gF_{t-1}$-measurable. By the definition of $q_{i,j}(S_\tau)$, we have
\begin{align*}
    \E\left[\sum_{\tau \le t, i \in S_\tau, j_\tau \le j}c_\tau(i,j) -
    q_{i,j}(S_\tau) \middle| \gF_{t-1} \right] = 0.
\end{align*}

	The number of $\tau$ satisfying $i \in S_\tau$ and $j_\tau \le j$ is exactly $SC_t(i,j) = \sum_{\tau=1}^t \mathbbm{1}[i \in S_\tau, j_\tau \le j]$. Therefore, by Azuma-Hoeffding inequality, we have for fixed $SC_t(i,j)$, with probability at least $1 - \delta$,
\begin{align*}
    \left|\sum_{\tau \le t, i \in S_\tau, j_\tau \le j} c_\tau(i, j) - \sum_{\tau \le t, i \in S_\tau, j_\tau \le j} q_{i,j}(S_\tau)\right| \le \sqrt{2SC_t(i,j)\log(T/\delta)},
\end{align*}
	By the union bound, we have
	\begin{align*}
	    \left|\sum_{\tau \le t, i \in S_\tau, j_\tau \le j} c_\tau(i, j) - \sum_{\tau \le t, i \in S_\tau, j_\tau \le j} q_{i,j}(S_\tau)\right| \le \sqrt{8SC_t(i,j)\log(NMT)},
\end{align*}
holds for any $t\in [T]$, $SC_t(i,j)$, and $(i, j) \in [N] \times [M]$ with probability at least $1 - T^{-2}$. We denote this good event as $\gE_1$ which satisfies $\Prob[\neg\gE_1] \le T^{-2}$.

We recall the definition of $\hat q_{i,j}^t$ given in \Cref{alg}:
\begin{align*}
    \hat q_{i,j}^t = \frac{C_t(i,j)}{SC_t(i,j)} = \frac{\sum_{\tau \le t} c_{\tau}(i,j)}{SC_t(i,j)} = \frac{\sum_{\tau \le t} \1[i \in S_\tau]\cdot c_{\tau}(i,j)\mathbbm{1}[j_\tau \le j]}{SC_t(i,j)}.
\end{align*}
Under this good event $\gE_1$, we have for every $t \in [T]$ and $(i, j) \in [N] \times [M]$,
\begin{align*}
	    \left| \hat q_{i,j}^t - \frac{\sum_{\tau \le t, i \in S_\tau, j_\tau \le j} q_{i,j}(S_\tau)}{SC_t(i,j)}\right| \le \sqrt{8\frac{\log(NMT)}{SC_t(i,j)}}.
\end{align*}

Below we bound the difference between $q^*_{i,j}(S_t)$ and $q_{i,j}(S_t)$ for any $S_t \in \gS$. For given $(i, j)$ with $i \in S_t$, we have
\begin{align*}
	    q^*_{i,j}(S_t) - q_{i,j}(S_t) &= \Prob[I_t \ni i, j_t = j \mid j_t \le j, S_t] - \Prob[i_t = i, j_t = j \mid j_t \le j, S_t] \\
    &= \Prob[I_t \ni i, i_t \neq i, j_t = j \mid j_t \le j, S_t] \\
    &\le \sum_{k \in S_t, k \neq i}\frac{\Prob[X_i \in M_j]\Prob[X_k \in M_j]}{\Prob[X_i \in M_{\le j}]\Prob[X_k \in M_{\le j}]} \\
    &\le (K-1)\cdot \frac{(L\epsilon)^2}{(j\epsilon/L)^2} = (K-1)\cdot L^4/j^2,
\end{align*}
where the last inequality holds by \Cref{ass:bi-lipschitz} and $\Prob[X_i \in M_{\le j}] = \sum_{j'=1}^j p_{i,j}^* \ge j\frac{\epsilon}{L}, \forall i \in [N]$.

Notice that for every $S_t \in \gS$ and $i \in S_t, j \in [M]$, we have
\begin{align*}
    q_{i,j}^*(S_t) &=
    \Prob[I_t \ni i, j = j_t \mid j_t \le j, S_t] \\
    &= \frac{\Prob[I_t \ni i, j = j_t \mid S_t]}{\Prob[j_t \le j \mid S_t]} \\
    &= \frac{\Prob[X_i(t) \in M_{j} \And x_k(t) \in M_{\le j}, \forall k \in S_t \mid S_t]}{\Prob[x_k(t) \in M_{\le j}, \forall k \in S_t \mid S_t]} \\
    &= \frac{\Prob[X_i \in M_j]\cdot \Prob[X_k \in M_{\le j}, \forall k \in S, k \neq i]}{\Prob[X_i \in M_{\le j}] \cdot \Prob[X_k \in M_{\le j}, \forall k \in S, k \neq i]} \\
    &= \frac{\Prob[X_i \in M_j]}{\Prob[X_i \in M_{\le j}]} \\
    &= \Prob[X_i \in M_j \mid X_i \in M_{\le j}]\\
    &= q_{i,j}^*.
\end{align*}
Therefore, we have
\begin{align*}
	    \left|\hat q_{i,j}^t - q^*_{i,j}\right| = \left|\hat q_{i,j}^t - \frac{\sum_{\tau \le t, i \in S_\tau, j_\tau \le j}q_{i,j}^*(S_\tau)}{SC_t(i,j)}\right| \le \sqrt{8\frac{\log(NMT)}{SC_t(i,j)}} + (K-1)\cdot(L^4/j^2).
\end{align*}
\end{proof}
\end{lemma}

\subsection{Optimistic Estimation}

\begin{lemma}
\label{lemma:optimism}
    For $\beta_{i,j}^t$ given in \Cref{eq:def-beta}, under event $\gE_0$ and $\gE_1$, we have
    \begin{align*}
        \bar q_{i,j}^t \ge q^*_{i,j}.
    \end{align*}
    Moreover, by the offline $(1-\epsilon)$-approximated optimization oracle PTAS \citep{chen2013combinatorial}, we have
    \begin{align*}
        \bar r_q(S_t, \bar \vq^t) \ge (1-\epsilon) \cdot \bar r_q(S^*; \bar \vq^t).
    \end{align*}
\begin{proof}
Notice that in \Cref{alg} we define
\begin{align*}
    \bar q_{i,j}^t = \min\left\{\hat q_{i,j}^t + \beta_{i,j}^t + \frac{(K-1)L^4}{j^2}, 1\right\}.
\end{align*}
By \Cref{lemma:concentration-formal}, we have under $\gE_0$ and $\gE_1$,
\begin{align*}
    \hat q_{i,j}^t \ge q_{i,j}^* - \beta_{i,j}^t - \frac{(K-1)L^4}{j^2},
\end{align*}
where the inequality holds by the definition of $\beta_{i,j}^t$ in \Cref{eq:def-beta} and $SC_{t-1}(i,j) \le SC_t(i,j)$.
Since $q_{i,j}^* \le 1$, we have
\begin{align*}
    \bar q_{i,j}^t \ge q^*_{i,j}.
\end{align*}

	Since in \Cref{alg}, we set action $S_t \leftarrow \operatorname{PTAS}(\bar \vp^t)$ where $\bar\vp^t$ is converted from $\bar\vq^t$ by \Cref{eq:qstar-def}. Then by \Cref{lemma:r-q-r,lemma:monotone}, we have
\begin{align*}
    \bar r_q(S_t; \bar\vq^t) = \bar r(S_t; \bar\vp^t)  \ge (1-\epsilon)\max_{S \in \gS} \bar r(S; \bar \vp^t) \ge (1-\epsilon)\bar r(S^*; \bar \vp^t) = (1-\epsilon) \bar r_q(S^*; \bar \vq^t).
\end{align*}
\end{proof}
\end{lemma}

\subsection{Regret Decomposition}

\begin{lemma}
\label{lemma:tpm-formal}
Denote $Q_j^*(S_t) :=  \prod_{k \in S_t, j' > j} (1 - q_{k,j'}^*)$. We have
\begin{align}
    |\bar r_q(S_t; \bar \vq^t) - \bar r_q(S_t; \vq^*)| \le 2\sum_{i \in S_t, j \in [M]} Q_j^*(S_t) \cdot v_j \cdot \left|\bar q_{i,j}^t - q^*_{i,j}\right|.
\end{align}
\begin{proof}
This lemma is given by directly applying Lemma 3.3 in \citet{wang2023combinatorial} by the definition of $\bar r_q$ in \Cref{eq:def-rq}.
\end{proof}
\end{lemma}

\begin{lemma}\label{lemma:regret-decomp}
Under \Cref{ass:bi-lipschitz}, we can bound the regret of \Cref{alg} by
\begin{align*}
    \gR(T) \le \E\left[\sum_{t=1}^T \texttt{Bonus}_t + \texttt{Bias}_t\middle| \gE_0, \gE_1\right] + 3T\epsilon + T^{-1},
\end{align*}
	where $\texttt{Bonus}_t$ and $\texttt{Bias}_t$ are defined by
\begin{align}\label{eq:def-bonus-t}
    \texttt{Bonus}_t := 4 \sum_{i \in S_t, j \in [M]} Q_{j}^{*}(S_t) \cdot v_j \cdot \beta_{i,j}^t,
\end{align}
and
\begin{align}\label{eq:def-bias-t}
    \texttt{Bias}_t := 4 \sum_{i \in S_t, j \in [M]} Q_{j}^{*}(S_t) \cdot v_j \cdot (K-1)\frac{L^4}{j^2}.
\end{align}
\begin{proof}
	This lemma formalizes the first three steps of the proof sketch. Denote $\Delta_t := r^*(S^*) - r^*(S_t)$, we have
	\begin{align*}
	    \gR(T) = \E\left[\sum_{t=1}^T \Delta_t\right].
\end{align*}
By \Cref{lemma:discrete-error}, we have
\begin{align*}
    \Delta_t &\le  \bar r(S^*; \vp^*) - \bar r(S_t; \vp^*) + 2\epsilon.
\end{align*}
Then we have
\begin{align*}
    \gR(T) &\le \Prob[\gE_0] \cdot \E\left[\sum_{t=1}^T\Delta_t \middle| \gE_0\right] + \Prob[\neg\gE_0] \cdot T \\
    &\le \E\left[\sum_{t=1}^T\Delta_t \middle| \gE_0\right] \\
    &\le \E\left[\sum_{t=1}^T\bar r(S^*; \vp^*) - \bar r(S_t; \vp^*) \middle| \gE_0\right] + 2T\epsilon,
\end{align*}
where the first inequality holds by property of conditional expectations and $\Delta_t \le 1$ and the second inequality is due to $\Prob[\neg\gE_0] = 0$.

Notice that under $\gE_0$ and $\gE_1$, by \Cref{lemma:monotone,lemma:optimism}, we have
\begin{align*}
    \bar r_q(S_t; \vq^t) \ge (1-\epsilon)\bar r_q(S^*; \bar \vq_t) \ge (1-\epsilon)\bar r_q(S^*; \vq^*).
\end{align*}
Then with \Cref{lemma:r-q-r-formal}, we have
\begin{align*}
    \gR(T) &\le \E\left[\sum_{t=1}^T\bar r_q(S^*; \vq^*) - \bar r_q(S_t; \vq^*) \middle| \gE_0\right] + 2T\epsilon \\
    &\le \E\left[\sum_{t=1}^T\bar r_q(S^*; \vq^*) - \bar r_q(S_t; \vq^*) \middle| \gE_0,\gE_1\right] + \Prob[\neg\gE_1]\cdot T +  2T\epsilon \\
    &\le \E\left[\bar r_q(S_t; \vq^t) - \bar r_q(S_t; \vq^*)\right] + 3T\epsilon + T^{-1},
\end{align*}
where the last inequality holds by $\epsilon\bar r_q(S^*;\vp^*) \le \epsilon$ and $\Prob[\neg\gE_1] \le T^{-2}$ shown in \Cref{lemma:concentration-formal}.

Therefore, applying \Cref{lemma:tpm-formal}, we get
\begin{align*}
    \gR(T) \le \E\left[\sum_{t=1}^T \texttt{Bonus}_t + \texttt{Bias}_t\middle| \gE_0, \gE_1\right] + 3T\epsilon + T^{-1},
\end{align*}
	where $\texttt{Bonus}_t$ and $\texttt{Bias}_t$ are defined in \Cref{eq:def-bonus-t,eq:def-bias-t}.
\end{proof}
\end{lemma}

\subsection{Bounding the Bonus Terms}

We apply similar methods in \citet{wang2017improving,liu2023contextual} to give the bounds of $\sum_t\texttt{Bonus}_t$. We first give the following definitions.

\begin{definition}[{\citet[Definition 5]{wang2017improving}}]\label{def:TPgroup}
    Let $(i,j) \in [N] \times [M]$ be the index of binary arm and $l$ be a positive natural number, define the triggering probability group (of actions)
    \[
    S_j^l = \{S \in \mathcal{S} \mid 2^{-l} < Q_j^*(S) \leq 2^{-l+1}\}.
    \]
    Notice $\{S_j^l\}_{l \geq 1}$ forms a partition of $\{S \in \mathcal{S} \mid Q_j^*(S) > 0\}$.
\end{definition}
\begin{definition}[{\citet[Definition 6]{wang2017improving}}]\label{def:TPcounter}
    For each group $S_j^l$ (\Cref{def:TPgroup}), we define a corresponding counter $N_{i,j}^l$.
    In a run of a learning algorithm, the counters are maintained in the following manner.
    All the counters are initialized to $0$. In each round $t$, if the action $S_t$ is chosen, then update $N^l(i,j)$ to $N^l(i,j) + 1$ for every $(i, j)$ that $i \in S_t$, $S_t \in S_j^l$.
    Denote $N_t^l({i,j})$ at the end of round $t$ with $N^l(i,j)$.
    In other words, we can define the counters with the recursive equation below:
    \begin{align*}
        N_t^l(i,j) =
        \begin{cases}
            0, & \text{if } t = 0, \\
            N_{t-1}^l(i,j) + 1, & \text{if } t > 0, i\in S_t, S_t \in S_j^l, \\
            N_{t-1}^l(i,j), & \text{otherwise}.
        \end{cases}
    \end{align*}
\end{definition}
\begin{definition}[{\citet[Definition 7]{wang2017improving}}]\label{def:TPevent}
Given a series of integers $\{l_{i,j}^{\max}\}_{i \in [N],j\in[M]}$, we say that the triggering is nice at the beginning of round $t$ (with respect to $l_{i,j}^{\max}$), if for every group $S_j^l$(\Cref{def:TPgroup}) identified by binary arm $(i,j)$ and $1 \leq l \leq l_{i,j}^{\max}$, as long as
\[
\sqrt{\frac{8 \log (NMT)}{\frac{1}{3} N_{t-1}^l(i,j)\cdot 2^{-l}}} \leq 1,
\]
there is $SC_{t-1}(i,j) \geq \frac{1}{3} N_{t-1}^l(i,j) \cdot 2^{-l}$. We denote this event with $\gE_2(t)$. It implies
\[
\beta_{i,j}^t = \sqrt{\frac{8 \log(NMT)}{ SC_{t-1}(i,j)}} \leq \sqrt{\frac{8 \log(NMT)}{\frac{1}{3} N_{t-1}^l(i,j) \cdot 2^{-l}}}.
\]
\end{definition}
Therefore, we show that $\gE_2(t)$ happens with high probability for every $t$.
\begin{lemma}[{\citet[Lemma 4]{wang2017improving}}]\label{lemma:TPprob}
For a series of integers $\{l_{i,j}^{\max}\}_{i \in [N],j\in[M]}$, $$\Prob[\neg \mathcal{E}_2(t)] \leq \sum_{i \in [N],j\in[M]} l_{i,j}^{\max} t^{-2},$$
for every round $t \geq 1$.

	\begin{proof}
We prove this lemma by showing $\Prob[N_{t-1}^l(i,j) = s, SC_{t-1}(i,j) \leq \frac{1}{3} N_{t-1}^l(i,j) \cdot 2^{-l}] \leq t^{-3}$, for any fixed $s$ with $0 \leq s \leq t - 1$ and $\sqrt{\frac{8 \log(NMT)}{\frac{1}{3} s \cdot 2^{-l}}} \leq 1$. Let $t_k$ be the round that $N^l(i,j)$ is increased for the $k$-th time, for $1 \leq k \leq s$. Let $Z_k = \1[S_{t_k} \ni i, j_{t_k} \le j]$ be a Bernoulli variable, that is, $SC_{t_k}(i,j)$ increase in round $t_k$. When fixing the action $S_{t_k}$, $Z_k$ is independent from $Z_1, \ldots, Z_{k-1}$. Since $S_{t_k} \in S_j^l$, $\mathbb{E}[Z_k \mid Z_1, \ldots, Z_{k-1}] \geq 2^{-l}$. Let $Z = Z_1 + \cdots + Z_s$. By multiplicative Chernoff bound \citep{upfal2005probability}, we have
\[
\Prob\left\{Z \leq \frac{1}{3} s \cdot 2^{-l}\right\} \leq \exp\left(-\frac{\left(\frac{2}{3}\right)^2 s \cdot 2^{-l}}{2}\right) \leq \exp\left(-\frac{\left(\frac{2}{3}\right)^2 18 \log t}{2}\right) < \exp(-3 \log t) = t^{-3}.
\]

By the definition of $SC_{t-1}(i,j)$ and the condition $N_{t-1}^l(i,j) = s$, we have $SC_{t-1}(i,j) \geq Z$. Thus
\begin{align*}
\Prob[N_{t-1}^l(i,j) = s, SC_{t-1}(i,j) &\leq \frac{1}{3} N_{t-1}^l(i,j) \cdot 2^{-l}]\\
&\leq \Prob[N_{t-1}^l(i,j) = s, Z \leq \frac{1}{3} s \cdot 2^{-l}] \\
&\leq \Prob[Z \leq \frac{1}{3} s \cdot 2^{-l}] \leq t^{-3}.
\end{align*}

	By taking $i,j$ over $[N]\times[M]$, $l$ over $1, \ldots, l_{i,j}^{\max}$, $s$ over $0, \ldots, t - 1$ and applying the union bound, the lemma holds.
\end{proof}
\end{lemma}

\begin{lemma}\label{lemma:bonus-t-bound}
For given constant $C$, we have
\begin{align*}
    \sum_{t=1}^T \texttt{Bonus}(t) \le 16NM + 12288\frac{KNM^2\log(NMT)}{C} + TC + \frac{\pi^2}{6} \left\lceil \log_2\frac{16KM}{C} \right\rceil.
\end{align*}
\begin{proof}
For given constant $C$, we can define the following notations.
\begin{align}\label{eq:def-lmax}
    l_{i,j}^{\max} := \left\lceil \log_2 \frac{16KM}{C} \right\rceil, \quad \forall (i, j) \in [N] \times [M],
\end{align}
and for every integer $l$,
\begin{equation}\label{eq:def-kappa}
\begin{aligned}
    \kappa_{l,T}(C,s) := \begin{cases}
        2\cdot 2^{-l} & s =0 \\
        \sqrt{{96 \cdot 2^{-l} \log(NMT)}/s} & 1 \le s \le B_{l,T}(C) \\
        0 & s > B_{l,T}(C)
    \end{cases},
\end{aligned}
\end{equation}
where $B_{l,T}(C)$ is given by
\begin{align}\label{eq:def-BlT}
    B_{l,T}(C) := \left\lfloor{6144 \cdot 2^{-l}K^2M^2\log(NMT)}/{C^2}\right\rfloor.
\end{align}
By \citet[Lemma 5]{wang2017improving}, if $\texttt{Bonus}(t) \ge C$, under event $\gE_2(t)$, we have
\begin{align*}
    \texttt{Bonus}(t) \le \sum_{i \in S_t, j \in [M]} \kappa_{l_{i,j}, T}(C, N_{t-1}^{l_i}(i,j)),
\end{align*}
where $l_{i,j}$ is the index of group $S_j^{l_{i,j}} \ni S_t$. This is because we have
\begin{align*}
    \texttt{Bonus}(t) &\le -C + 8 \sum_{i \in S_t, j \in [M]} Q_j^*(S_t) \cdot (j - 1)\epsilon \cdot \min\{\beta_{i,j}^t, 1\} \\
    &\le 8 \sum_{i \in S_t, j \in [M]} \left(Q_j^*(S_t) \cdot \min\{\beta_{i,j}^t, 1\} - \frac{C}{8KM}\right)
\end{align*}

\noindent \textbf{Case 1: $1\le l_{i,j} \le l_{i,j}^{\max}$.} We have
\begin{align*}
    Q^*_{j}(S_t) \le 2 \cdot 2^{-l_{i,j}}.
\end{align*}
Under $\gE_2(t)$, we have
\begin{align*}
    \min \left\{\beta_{i,j}^t, 1\right\} = \min \left\{\sqrt{\frac{8 \log(NMT)}{ SC_{t-1}(i,j)}},1\right\} \leq \min\left\{\sqrt{\frac{8 \log(NMT)}{\frac{1}{3} N_{t-1}^{l_{i,j}}(i,j) \cdot 2^{-l_{i,j}}}}, 1\right\},
\end{align*}
and
\begin{equation}\label{eq:Qbeta-bound}
\begin{aligned}
    Q^*_{j}(S_t)\cdot \min \left\{\beta_{i,j}^t,1\right\} &\le 2 \cdot 2^{-l_{i,j}} \cdot \min\left\{\sqrt{\frac{8 \log(NMT)}{\frac{1}{3} N_{t-1}^{l_{i,j}}(i,j) \cdot 2^{-l_{i,j}}}}, 1\right\} \\
    &\le \min\left\{\sqrt{\frac{96 \cdot 2^{-l_{i,j}} \log(NMT)}{ N_{t-1}^{l_{i,j}}(i,j) }}, 2 \cdot 2^{-l_{i,j}}\right\}.
\end{aligned}
\end{equation}
If $N_{t-1}^{l_{i,j}}(i,j) \ge B_{l_{i,j},T}(C) + 1$, then
\begin{align*}
    \sqrt{\frac{96 \cdot 2^{-l_{i,j}} \log(NMT)}{ N_{t-1}^{l_{i,j}}(i,j) }} \le \frac{C}{8KM},
\end{align*}
which implies $Q^*_{j}(S_t)\cdot \min \left\{\beta_{i,j}^t,1\right\} - C/8KM \le 0$.

If $N_{t-1}^{l_{i,j}}(i,j) = 0$, we have $Q^*_{j}(S_t)\cdot \min \left\{\beta_{i,j}^t,1\right\} \le Q^*_{j}(S_t) \le 2\cdot 2^{-l_{i,j}}$, which implies
\begin{align*}
    Q^*_{j}(S_t)\cdot \min \left\{\beta_{i,j}^t,1\right\} - \frac{C}{8KM} \le \kappa_{l_{i,j},T}(C, 0)
\end{align*}

Otherwise, for $1 \le N_{t-1}^{l_{i,j}}(i,j) \le B_{l_{i,j}, T}(C)$, we have $Q^*_{j}(S_t)\cdot \min \left\{\beta_{i,j}^t,1\right\} \le \kappa_{l_{i,j},T}(C, N_{t-1}^{l_{i,j}}(i,j))$ by \Cref{eq:Qbeta-bound,eq:def-kappa}. Therefore, we get
\begin{align*}
    Q^*_{j}(S_t)\cdot \min \left\{\beta_{i,j}^t,1\right\} - \frac{C}{8KM} \le \kappa_{l_{i,j},T}(C, N_{t-1}^{l_{i,j}}(i,j))
\end{align*}

\noindent \textbf{Case 2: $l_{i,j} \ge l_{i,j}^{\max} + 1$.} We have
\begin{align*}
    Q^*_{j}(S_t)\cdot \min \left\{\beta_{i,j}^t,1\right\} \le 2 \cdot 2^{-l_{i,j}} \le 2 \cdot \frac{C}{16KM} \le \frac{C}{8KM},
\end{align*}
which shows that $Q^*_{j}(S_t)\cdot \min \left\{\beta_{i,j}^t,1\right\} - C/8KM \le 0$. If $N_{t-1}^{l_{i,j}}(i,j) = 0$. Therefore, we finally get
\begin{align*}
    \texttt{Bonus}(t) \le 8\sum_{i\in S_t, j \in [M]} \kappa_{l_{i,j}, T}\left(C, N_{t-1}^{l_{i,j}}(i,j)\right),
\end{align*}
for the case of good event $\gE_2(t)$ happens and $\texttt{Bonus}(t) \ge C$.

Notice that under good events $\gE_0, \gE_1$, we have
\begin{align*}
    \sum_{t=1}^T \texttt{Bonus}(t) &\le \sum_{t=1}^T \1[\{\texttt{Bonus}(t) \ge C\} \cap \gE_2(t)]\cdot \texttt{Bonus}(t) + T \cdot C + \sum_{t=1}^T \Prob[\gE_2(t)] \\
    &\le \underbrace{\sum_{t=1}^T 8 \cdot \sum_{i\in S_t, j \in [M]} \kappa_{l_{i,j}, T}\left(C, N_{t-1}^{l_{i,j}}(i,j)\right)}_{(I)} + TC + \frac{\pi^2}{6} \cdot \max_{i\in [N], j \in [M]} l_{i,j}^{\max} .
\end{align*}
where the first inequality is due to $\texttt{Bonus}(t) \le 1$ and definition, and the second one is due to \Cref{lemma:TPprob}. The key is bounding $(I)$:
\begin{align*}
    (I) &= 8\cdot \sum_{i \in [N], j \in [M]} \sum_{l=1}^\infty \sum_{s=0}^{N_{T-1}^{l}(i,j)}\kappa_{l}(C, s) \\
    &= 8 \cdot \sum_{i \in [N], j \in [M]} \sum_{l=1}^\infty \left(2\cdot 2^{-l} +  \sum_{s=1}^{B_{l,T}(C)}\sqrt{\frac{96 \cdot 2^{-l} \log(NMT)}{ s }} \right) \\
    &\le 8\cdot  \sum_{i \in [N], j \in [M]} \sum_{l=1}^\infty \left(2\cdot 2^{-l} +  2\cdot \sqrt{96 \cdot 2^{-l} \log(NMT)}\cdot \sqrt{B_{l,T}(C)} \right) ,
\end{align*}
where the inequality holds by the fact that $\sum_{s=1}^n \sqrt{1/s} \le 2\sqrt{n}$. Therefore, by the definition of $B_{l,T}(C)$ in \Cref{eq:def-BlT}, we have
\begin{align*}
    (I) &\le  8\cdot  \sum_{i \in [N], j \in [M]} \sum_{l=1}^\infty \left(2\cdot 2^{-l} +  1536\cdot \frac{2^{-l}KM\log(NMT)}{C} \right) \\
    &= 8\cdot  \sum_{i \in [N], j \in [M]}  \left(2+  1536\cdot \frac{KM\log(NMT)}{C} \right)\cdot \left(\sum_{l=1}^\infty 2^{-l}\right) \\
    &\le 16NM + 12288\frac{KNM^2\log(NMT)}{C}.
\end{align*}
Therefore, we get
\begin{align*}
    \sum_{t=1}^T \texttt{Bonus}(t) \le 16NM + 12288\frac{KNM^2\log(NMT)}{C} + TC + \frac{\pi^2}{6} \left\lceil \log_2\frac{16KM}{C} \right\rceil.
\end{align*}
\end{proof}
\end{lemma}

\subsection{Bounding the Bias Terms}
\begin{lemma}\label{lemma:bias-t-bound}
Under \Cref{ass:bi-lipschitz}, we have
\begin{align*}
    \sum_{t=1}^T \texttt{Bias}(t) \le 4K^2L^4 T\epsilon \log(M + 1).
\end{align*}
\begin{proof}
Notice that $\sum_{j\in[M]} 1/j \le \log(M+1)$ for $\epsilon < 1/2$, we have
\begin{align*}
    \texttt{Bias}(t) &\le 4K \cdot \sum_{i \in S_t,j \in [M]}Q^*_{j}(S_t) \cdot \epsilon L^4/j \\
    &= 4K^2L^4\epsilon \cdot \sum_{j \in [M]} \frac{1}{j} \\
    &\le 4K^2L^4 \epsilon\log(M + 1).
\end{align*}
Therefore, we have
\begin{align*}
    \sum_{t=1}^T \texttt{Bias}(t) \le 4K^2L^4 T\epsilon \log(M+1).
\end{align*}
\end{proof}
\end{lemma}

\subsection{Proof of Theorem~\ref{thm:main}}
\begin{theorem}[Formal version of \Cref{thm:main}]\label{thm:main-formal}
By setting $\beta_{i,j}^t$ in \Cref{eq:def-beta} and $\epsilon < 1/2$, we can control the regret of \Cref{alg} under \Cref{ass:bi-lipschitz} by
\begin{align*}
	    \gR(T) &\le  12289 \sqrt{NKM^2T\log(NMT)} +  T\epsilon\left(4K^2L^4\log(M+1) + 3\right)\\
    &\quad + 16 NM +  \pi^2\left(\log_2(\sqrt{KM^2T\log(NMT)/N}) + 5\right)/6 +  T^{-1} \\
    &= \wt{O}\left(\sqrt{NKM^2T} + L^4K^2T\epsilon\right),
\end{align*}
where $M = \ceil{1/\epsilon}$. If we further take $\epsilon = \gO\left(L^{-2}K^{-\frac{3}{4}}N^{\frac
{1}{4}}T^{-\frac{1}{4}}\right)$, we have
\begin{align*}
    \gR(T) = \wt{\gO}(L^{2}N^{\frac{1}{4}}K^{\frac{5}{4}}T^{\frac{3}{4}}).
\end{align*}
\begin{proof}
By \Cref{lemma:regret-decomp}, we have
\begin{align*}
    \gR(T) \le \E\left[\sum_{t=1}^T \texttt{Bonus}_t + \texttt{Bias}_t\middle| \gE_0, \gE_1\right] + 3T\epsilon + T^{-1},
\end{align*}
Take constant $C$ as
\begin{align}\label{eq:def-C}
    C:= \sqrt{\frac{NKM^2\log(NMT)}{T}}.
\end{align}
Then \Cref{lemma:bonus-t-bound} shows that
\begin{align*}
    \sum_{t=1}^T \texttt{Bonus}(t) \le 16NM + 12289\sqrt{NM^2KT\log(NMT)} + \pi^2\left(\log_2(\sqrt{KMT\log(NMT)/N}) + 5\right)/6.
\end{align*}
\Cref{lemma:bias-t-bound} demonstrates that
\begin{align*}
    \sum_{t=1}^T \texttt{Bias}(t) \le 4K^2L^4 T\epsilon \log(M + 1).
\end{align*}
Therefore, by calculating the summation of the bonus and bias terms, we can bound the regret by
\begin{align*}
    \gR(T) &\le \E\left[\sum_{t=1}^T \texttt{Bonus}(t) + \texttt{Bias}(t) \middle| \gE_0, \gE_1\right] +  T^{-1} + 3T\epsilon \\
	    &\le  12289 \sqrt{NKM^2T\log(NMT)} +  T\epsilon\left(4K^2L^4\log(M+1) + 3\right)\\
    &\quad + 16 NM +  \pi^2\left(\log_2(\sqrt{KM^2T\log(NMT)/N}) + 5\right)/6 +  T^{-1} \\
    &= \wt{O}\left(\sqrt{NKM^2T} + L^4K^2T\epsilon\right),
\end{align*}
which finishes the proof.
\end{proof}
\end{theorem}

\section{Numerical Experiments for \texttt{DCK-UCB}}\label{app:experiments}

We conduct numerical experiments to validate the effectiveness of our proposed \texttt{DCK-UCB} algorithm.

\subsection{Comparable Baselines}
We compare \texttt{DCK-UCB} with two baseline algorithms:
\textbf{(1) Naive UCB}: A naive adaptation of UCB \cite{lattimore2020bandit} that treats each subset of $K$ base arms (i.e., each super arm) as an abstract arm, with $\binom{N}{K}$ super arms in total.
\textbf{(2) Submodular Greedy}: A submodularity-driven greedy algorithm (widely used for submodular maximization task \cite{nie2022explore,tajdini2024nearly}) that exploits the submodularity of $K$-Max bandits, i.e., it repeats the process that uniformly explores available base arms and greedily includes the best one as part of its solution, until the solution is of size $K$.

We highlight that standard baselines for $K$-Max bandits, such as SDCB \citep{chen2016combinatorial} and CUCB \citep{wang2023combinatorial}, are not suitable for continuous $K$-Max bandits with value-index feedback. Specifically, SDCB \citep{chen2016combinatorial} requires semi-bandit feedback, which demands observations from every selected arm. In our case, the agent receives only the maximum outcome with the winner's index. CUCB \citep{wang2023combinatorial} works in finite-support distribution settings by recording and estimating each observed outcome value. For continuous $K$-Max bandits, this would require unbounded memory and is therefore not directly applicable.

\subsection{Instance Setup}
We first construct the instance with $(N = 10, K = 5)$ and $(N = 12, K = 3)$, where each arm's outcome is piecewise uniform in $[0, 1]$ with bi-Lipschitz constant $L = 3$.

\paragraph{Instance generation.}
All synthetic instances are built from independent \emph{piecewise-uniform} base arms on $[0,1]$.
This family gives a controlled continuous testbed: the bounded densities satisfy the bi-Lipschitz condition in \Cref{ass:bi-lipschitz}, while random breakpoints and segment heights create heterogeneous, nonparametric shapes beyond a single simple parametric family.
Piecewise-constant densities also make the ground-truth rewards $\EE[\max_{i\in S}X_i]$ and optimal subsets numerically stable to compute, reducing benchmark-construction noise in the reported regret.
For given $(N,A,K,L)$, we construct $N$ base arms $X_1,\dots,X_N$ as follows.
For each arm $i$, we first draw the number of breakpoints $B_i$ uniformly from $\{\lfloor K/2\rfloor\!+\!1,\dots,K\}$ and sample $B_i$ i.i.d. points in $(0,1)$, which are sorted into $0=b_{i,0}<b_{i,1}<\cdots<b_{i,B_i}<b_{i,B_i+1}=1$.
On each segment $[b_{i,j},b_{i,j+1})$ we draw a density level $p_{i,j}\sim\mathrm{Unif}[1/L,L]$ and normalize $\{p_{i,j}\}$ so that the piecewise-constant pdf integrates to $1$.
Hence each arm has a piecewise-linear cdf $F_i$ and a piecewise-constant pdf (implementation in our generator).

\begin{figure}[t]
  \centering
  \begin{minipage}{0.485\textwidth}
    \centering
    \textbf{(a) $N=10, K=5$}\par
    \vspace{2pt}
    \includegraphics[width=\linewidth]{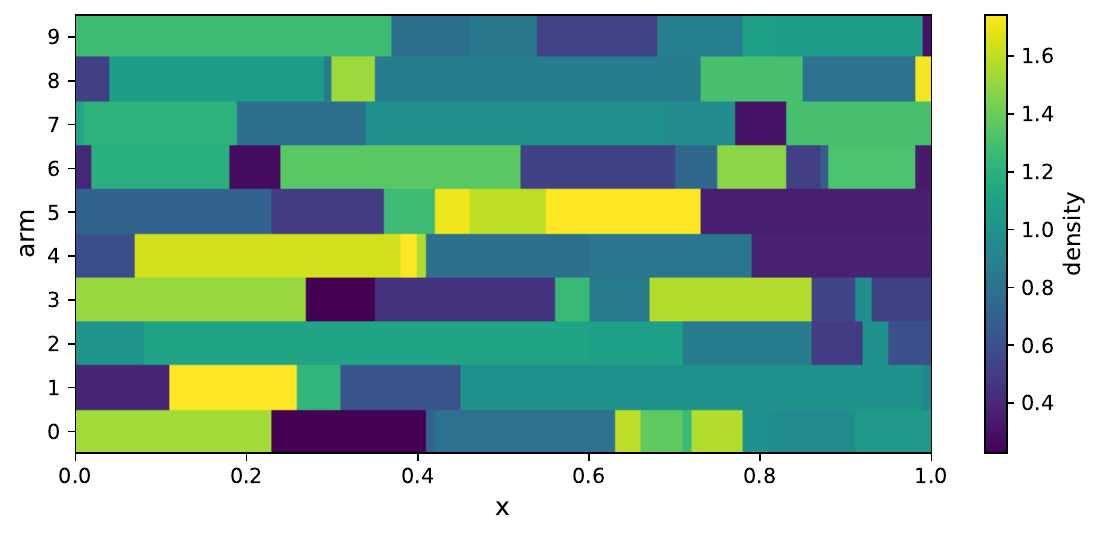}
  \end{minipage}\hfill
  \begin{minipage}{0.485\textwidth}
    \centering
    \textbf{(b) $N=12, K=3$}\par
    \vspace{2pt}
    \includegraphics[width=\linewidth]{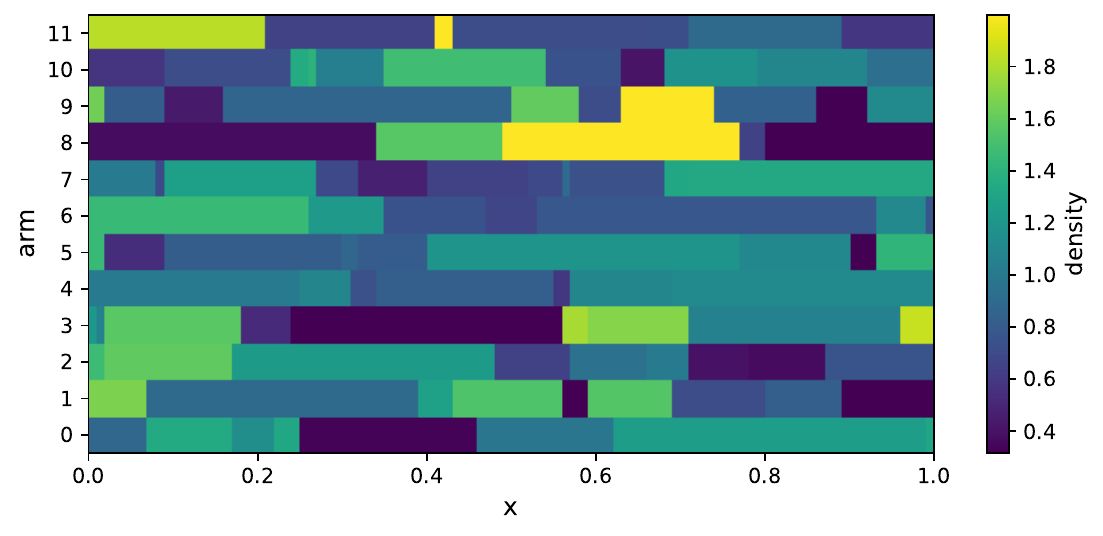}
  \end{minipage}
  \caption{Per-arm density heatmaps for two benchmark instances.
  Rows correspond to arm indices and the horizontal axis is $x\in[0,1]$; color encodes the probability density.
  Brighter cells indicate higher mass near that $x$.}
  \label{fig:instance-heatmaps}
\end{figure}

\paragraph{Avoid Trivial Cases.}
Our greedy baseline adds one arm at a time by maximizing the marginal gain in $\EE[\max_{i\in S}X_i]$ (ties broken by smaller index).
To avoid trivial cases, we \emph{reject} any randomly drawn instance unless this greedy rule is suboptimal:
\[
\EE\!\left[\max_{i\in S^\star}X_i\right] - \EE\!\left[\max_{i\in S_{\mathrm{greedy}}}X_i\right] \;\ge\; \texttt{gap},
\]
where \texttt{gap} is a user-specified threshold.
Concretely, we repeatedly resample the $N$ arms (up to a fixed maximum number of attempts) until the above inequality holds, at which point the instance is accepted.
Figure~\ref{fig:instance-heatmaps} summarizes the two instances used in our experiments via per-arm density heatmaps.

Code and instance generator are provided in the supplementary material.

\subsection{Experiment Results}

\begin{figure}[t]
    \centering
    \begin{minipage}{0.49\textwidth}
        \centering
        \includegraphics[width=\textwidth]{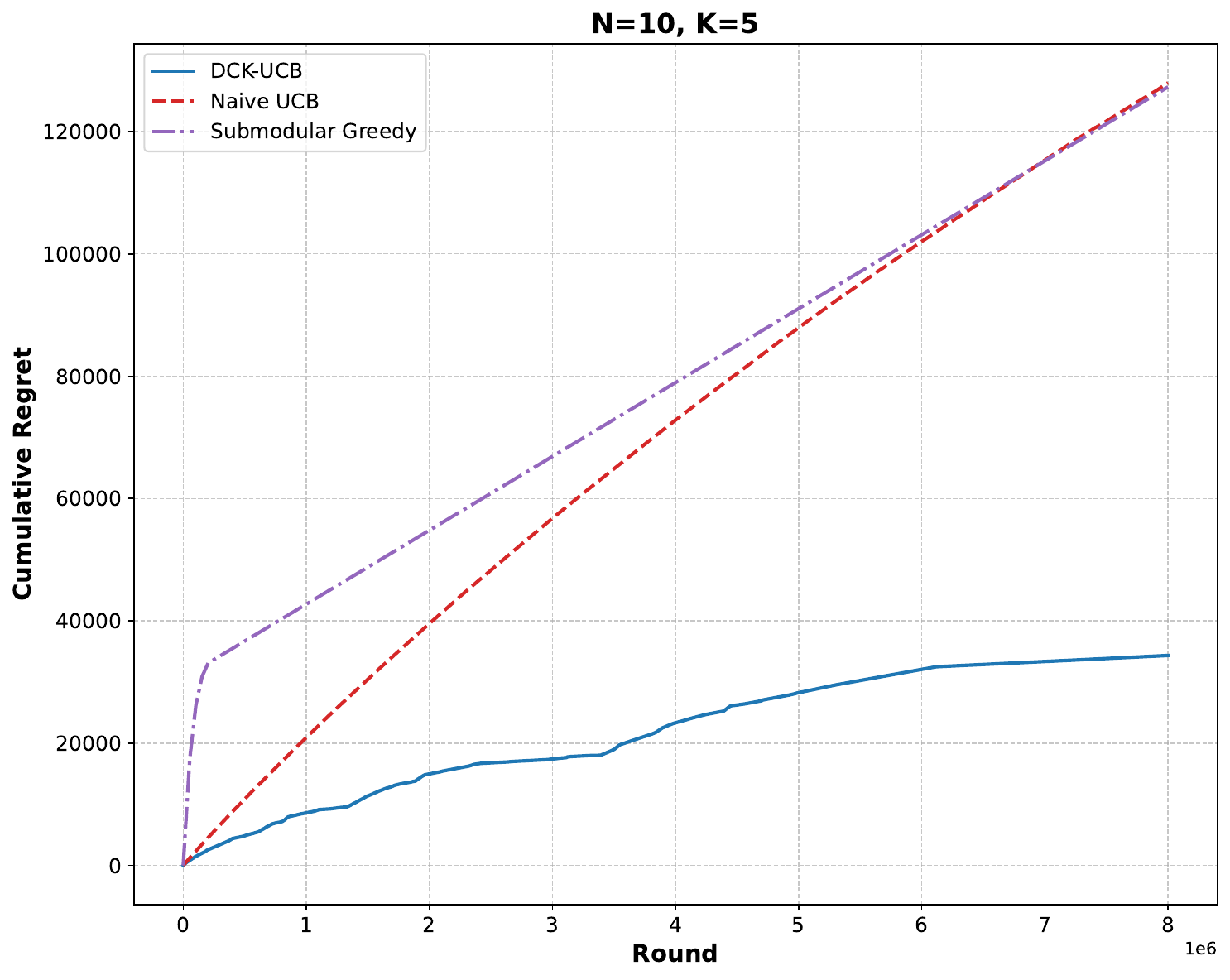}
        \caption{Comparison of cumulative regret for different algorithms on the problem instance with $N=10$ and $K=5$.}
        \label{fig:exp_comparison_N10A5}
    \end{minipage}
    \hfill
    \begin{minipage}{0.49\textwidth}
        \centering
        \includegraphics[width=\textwidth]{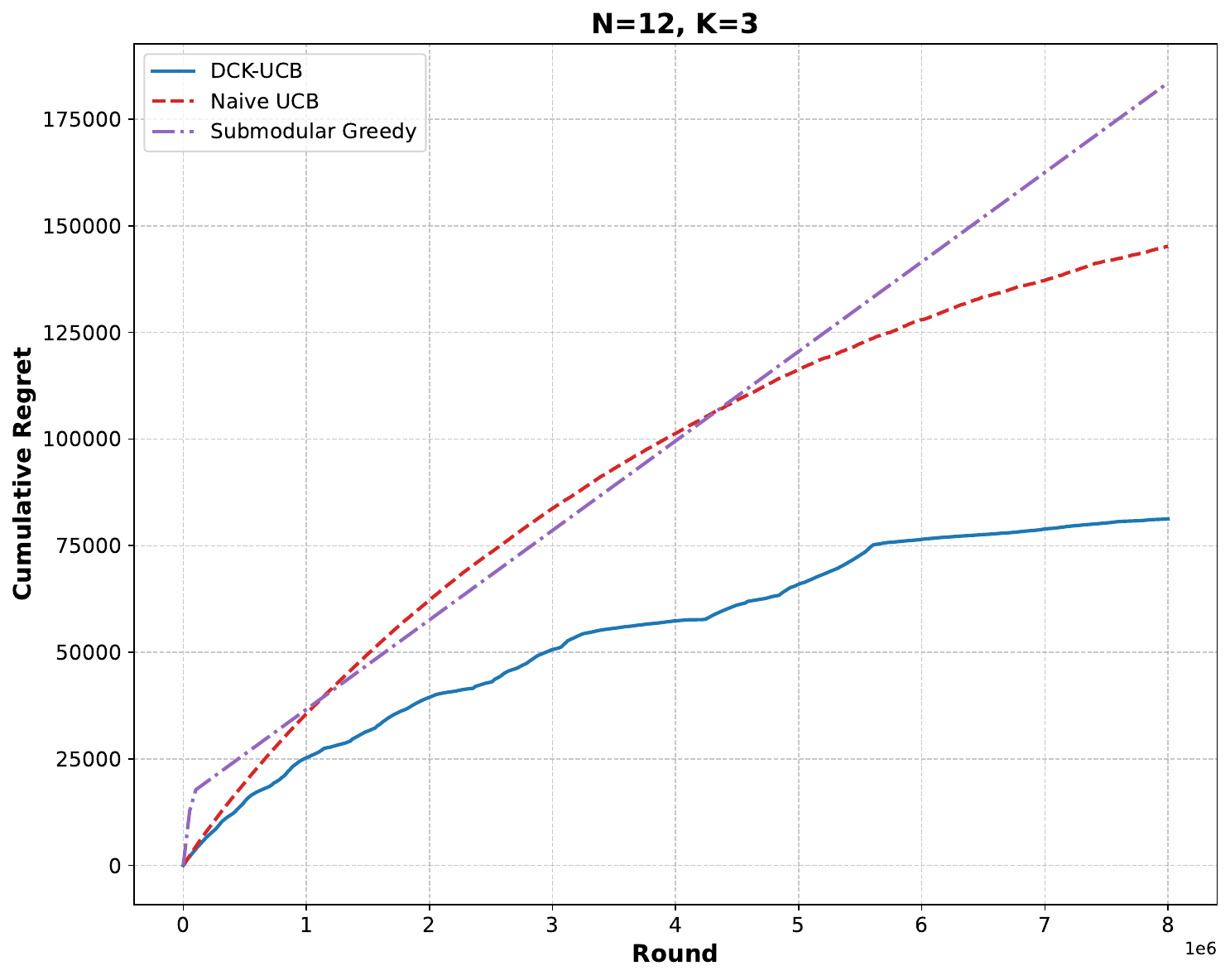}
        \caption{Comparison of cumulative regret for different algorithms on the problem instance with $N=12$ and $K=3$.}
        \label{fig:exp_comparison_N12A3}
    \end{minipage}
\end{figure}

From Figures \ref{fig:exp_comparison_N10A5} and \ref{fig:exp_comparison_N12A3}, we can observe that our proposed \texttt{DCK-UCB} algorithm consistently outperforms the baseline algorithms across different problem settings. In Figure \ref{fig:exp_comparison_N10A5}, where we have $N=10$ arms and need to select $K=5$ arms in each round, \texttt{DCK-UCB} achieves significantly lower cumulative regret compared to both the Submodular Greedy and Naive UCB approaches.

The linear growth in regret observed for the Submodular Greedy algorithm stems from evaluating regret against the true optimal subset rather than an approximation benchmark. In our experimental setup, the optimal action consistently outperforms the greedy selection, creating a persistent gap that accumulates linearly over time.

The Naive UCB approach performs particularly poorly in this setting, likely due to the large number of super arms ($\binom{10}{5} = 252$ in total).

Similarly, in Figure \ref{fig:exp_comparison_N12A3}, which presents a different configuration with $N=12$ arms and $K=3$, our algorithm maintains its strong performance. The Naive UCB approach performs better in this setting compared to the previous configuration, exhibiting a sublinear growth in regret with respect to round $t$. This improved performance can be attributed to the smaller number of super arms ($\binom{12}{3} = 220$ super arms compared to $\binom{10}{5} = 252$ in the previous configuration). Nevertheless, \texttt{DCK-UCB} still achieves lower cumulative regret.

These experimental results support our theoretical findings and demonstrate that \texttt{DCK-UCB} effectively balances the exploration-exploitation trade-off while handling the continuous nature of the arm distributions and the combinatorial structure of base arms.

\section{Detailed Explanation on \texttt{MLE-Exp} for Exponential K-Min Bandits}
\label{app:kminexp}
\subsection{Motivation}

Exponential distributions naturally model arrival or failure times in networked systems, job completion times in distributed computing, and service durations in queuing systems.
A canonical application arises in server scheduling, where the goal is to select $K$ servers to minimize the service latency. Here, each server's latency can be modeled as an exponential random variable with a rate parameter $\mu_i$, and the overall performance of the $K$ selected servers is the lowest latency achieved among them.
Here, the random outcome $X_i$ can be viewed as a random loss, and the winning loss is the minimum one.
Moreover, we consider a linear parameterization to parameter $\mu_i$, which allows incorporating features like distance, traffic, or weather conditions into the model.

\subsection{The K-Min Exponential Bandits}

Motivated by this, we consider a special case of $K$-Max bandits: the $K$-Min exponential bandits.
Here each arm $i$ generates loss $X_i$ from an exponential distribution with linear parameterization.
Specifically, each outcome distribution is an exponential distribution, i.e., $X_i \sim D_i = \Exp(\mu_i)$ where $\mu_i > 0$ is the parameter of arm $i$. Moreover, we assume that there exists a $d$-dimensional unknown parameter $\theta^* \in \R^d$ and a known feature mapping $\phi : [N] \to \R^d$ such that $\mu_i = \langle \phi(i), \theta^* \rangle$ holds for any $i \in [N]$. The feature mapping $\phi$ satisfies that $\|\phi(i)\|_2 \le 1$ and the unknown parameter $\theta^*$ satisfies $\theta^* \in \Theta  \subset \R^d$, where $\sup_{\theta \in \Theta} \|\theta\|_2 \le V$.
The agent observes \textit{only} the minimum loss $\ell_t = \min_{i \in S_t} X_i(t)$ after playing subset $S_t \in \mathcal{S} = \{S \subseteq [N] : |S| = K\}$.
That is, we consider the weaker full bandit feedback case.

Let $\ell^*(S) := \E[\ell_t \mid S]$ be the expected loss for action $S \in \gS$, we further denote the best action $S^* = \argmin_{S \in \gS} \ell^*(S)$ and similarly introduce the regret metric to evaluate the performance of this agent:
\begin{align*}
    \gR(T) = \E\left[\sum_{t=1}^T \ell^*(S_t) - \ell^*(S^*)\right].
\end{align*}

Note that we can let $Z_i(t)= - X_i(t)$ and view $Z_i(t)$ as a kind of reward, and let $r_t = \max_{i \in S_t} Z_i(t)$. Then $\ell_t = \min_{i \in S_t} X_i(t) = \min_{i \in S_t} -Z_i(t) = - \max_{i \in S_t} Z_i(t) = -r_t$. Thus, we can view $K$-Min exponential bandits as a special case of $K$-Max bandits.
However, one important difference is that in $K$-Min exponential bandits, we do not have value-index feedback, i.e., we do not know the winner's index. This is a full \textit{bandit feedback} setting, which makes $K$-Min exponential bandits even more challenging.

\subsection{Algorithm}

The key observation in $K$-Min exponential bandits is that the minimum of several exponential distributions still follows an exponential distribution. That is, we have
\begin{align*}
    \min_{i \in S} X_i \sim \Exp\left(\sum_{i \in S} \mu_i\right) = \Exp\left(\sum_{i \in S} \langle \phi(i), \theta^*\rangle \right).
\end{align*}
Therefore, it becomes much easier to estimate the true parameter $\theta^*$ by MLE.
Specifically, let $\psi(S)  := \sum_{i \in S} \phi(i)$, $\forall S \in \gS$. Then with chosen action $S_t$ and parameter $\theta$, the observed loss should follow the exponential distribution $\Exp\left(\sum_{i \in S} \phi(i)^T \theta \right) = \Exp\left(\psi(S)^T\theta\right)$, whose probability density function is $f(x) = (\psi(S)^T\theta) \cdot e^{\left(-(\psi(S)^T\theta)  x\right)}$.
Because of this, the log-likelihood function is
\begin{equation}
\begin{aligned}
    L_t(\ell_t; S_t, \theta) :&= \log \left( \psi(S_t)^\top \theta e^{\left( -\psi(S_t)^\top \theta \ell_t\right)} \right).
\end{aligned}
\end{equation}

Denote $\gL_t(\theta)$ as the summation of $L_t$ and a regularization term
\begin{align}\label{eq:gL-app}
    \gL_t(\theta; \lambda) := \sum_{i < t} - L_i(\ell_i; S_i, \theta) + \frac{\lambda}{2}\|\theta\|^2,
\end{align}
where $\lambda$ is the regularization factor. Then we present the algorithm \texttt{MLE-Exp} for $K$-Min exponential bandits in \Cref{alg:k-min-app}.

\begin{algorithm}
\caption{\texttt{MLE-Exp}: MLE for $K$-Min Exponential Bandits}
\label{alg:k-min-app}
\begin{algorithmic}[1]
\REQUIRE Regularization factors $\{\lambda_t\}_{t\in [T]}$, confidence radius $\{\gamma_t\}_{t\in [T]}$, and probability constant $\delta$.
\FOR{$t = 1, \ldots, T$}
    \STATE Compute MLE $\hat\theta_t$ by
    $$\hat{\theta}_t \leftarrow \argmin_{\theta \in \R^d} \mathcal{L}_t(\theta; \lambda_t),$$
    where $\gL_t(\theta;\lambda)$ is given in \Cref{eq:gL-app}.
	    \STATE Construct the confidence set $C_t(\hat\theta_t; \delta, \lambda_t)$ according to \Cref{eq:def-confidence-set}
	    \STATE $(S_t, \wt{\theta}_t) \leftarrow \argmax_{S \in \gS, \theta \in C_t(\hat{\theta}_t; \delta, \lambda_t)} \langle \psi(S), \theta \rangle$
	    \STATE Play action $S_t$ and observe the loss $\ell_t$.
\ENDFOR
\end{algorithmic}
\end{algorithm}

In Line 2 of \Cref{alg:k-min-app}, we estimate the MLE $\hat\theta_t$ by minimizing the summation of the log-likelihood function and the regularization term $\gL_t(\theta, \lambda_t)$.
Given $\lambda_t$ a priori, we will write $\gL_t(\theta)$ instead of $\gL_t(\theta, \lambda_t)$ for simplicity.
Inspired by \citet{liu2024almost, lee2024unified, liu2024combinatorial}, in Line 3, we construct a confidence set $C_t(\hat\theta_t;\delta)$ (defined in \Cref{eq:def-confidence-set}), centered at the MLE $\hat\theta_t$ with confidence radius $\gamma_t(\delta)$, based on the gradient term $g_t(\theta) := -\nabla_\theta \gL_t(\theta) + \sum_{i < t} \ell_i \psi(S_i)$ and Hessian matrix $H_t(\theta) := \nabla^2_\theta \gL_t(\theta)$.
Then in Line 4, we apply a double oracle to look for the action $S$ whose expected loss under a parameter $\theta$ in the confidence set ($1/\langle \psi(S), \theta \rangle$) is minimized.
Finally, we select this greedy action in Line 5 and use the observation to update the next time step's MLE and confidence set.

\Cref{thm:kminexp-formal} gives a formal version of \Cref{thm:kminexp}, which achieves the first nearly-minimax optimal regret bound $\wt{\gO}(\sqrt{T})$ for $K$-Min exponential bandits. This result advances the general result \Cref{thm:main} in three aspects.

\textbf{(1)} We utilize the property of exponential distributions in $K$-Min bandits and achieve efficient unbiased estimation of the true parameter $\theta^*$ without the index feedback required in \Cref{alg} for general continuous distributions.

\textbf{(2)} We consider maximum log-likelihood estimator $\hat\theta_t$ and achieve the following concentration lemma inspired by the analysis of general linear bandits \citep{lee2024unified,liu2024almost} and logistics bandits \citep{liu2024combinatorial}.
\begin{lemma}
   With probability at least $1-\delta$, set $\lambda_t$ in \Cref{eq:def-lambda} and $\gamma_t$ in \Cref{eq:def-gamma}, we have $\theta^* \in C_t(\hat\theta_t; \delta)$ holds for every $t \in [T]$.
\end{lemma}
Equipped with the efficient MLE, we sharpen the cumulative regret bound from $\wt{\gO}(T^{3/4})$ in \Cref{thm:main} to $\wt{\gO}(\sqrt{T})$.

\textbf{(3)} Thanks to the unbiased estimation of $\theta^*$, we analyze the regret bound with the Hessian matrix $H_t(\theta)$ and avoid the inverse Lipschitz factor in \Cref{ass:bi-lipschitz}, which might be infinite for exponential distributions.

\section{Omitted proofs in Section~\ref{sec:kminexp}}\label{Appendix:k-min}
This proof mainly applies the techniques for general linear bandits \citet{liu2024almost, lee2024unified}. Given action $S \in \gS$ to the environment, we assume that $\ell_S$ is the random variable of the loss, i.e., $\E[\ell_S] = \ell^*(S)$.

We have
\begin{align*}
    g_t(\theta; \lambda) :&= - \nabla_\theta \gL_t(\theta; \lambda) + \sum_{i < t} \ell_i \psi(S_i) \\
    &=  \sum_{i < t}  \frac{1}{\psi(S_i)^\top \theta} \cdot \psi(S_i) - \lambda \theta. \\
	    H_t(\theta;\lambda) :&= \nabla^2_\theta \gL_t(\theta; \lambda)\\
    &=  - \nabla_\theta g_t(\theta; \lambda)  \\
    &= \lambda I + \sum_{i < t} \frac{\psi(S_i)\psi(S_i)^\top}{(\psi(S_i)^\top\theta)^2}.
\end{align*}
\subsection{Concentration Argument for MLE}
\begin{lemma}[MLE Concentration]
\label{lemma:mle-concentration}
For $L^* := \sup_{S \in \gS} \ell^*(S)$, $M_1:= 2L^*$, and $V = \sup\{\|\theta\|_2 : \theta \in \Theta\}$, set
\begin{align}\label{eq:def-lambda}
    \lambda_t := \max \left\{1, \frac{2dM_1}{V}\cdot \log\left(e\sqrt{1 + \frac{tL^*}{d}} + \frac{1}{\delta}\right)\right\},
\end{align}
and
\begin{align} \label{eq:def-gamma}
    \gamma_t(\delta, \lambda_t) := \sqrt{\lambda_t}\left(\frac{1}{2M_1} + V\right) + \frac{2M_1d}{\sqrt{\lambda_t}}\left(\log(2) + \frac{1}{2}\log\left(1 + \frac{tL^*}{\lambda_t d}\right)\right) + \frac{2M_1}{\sqrt{\lambda_t}}\log(1/\delta).
\end{align}
Then we have with probability at least $1 - \delta$,
\begin{align}\label{eq:def-confidence-set}
    \theta^* \in C_t(\hat\theta_t; \delta,\lambda_t) := \left\{ \theta \in \Theta : \left\| g_t(\theta; \lambda_t) - g_t(\hat\theta_t; \lambda_t) \right\|_{H_t^{-1}(\theta; \lambda_t)} \le \gamma_t(\delta, \lambda_t) \right\},
\end{align}
holds for any $t \in [T]$. We denote the confidence set as $C_t(\hat\theta_t; \delta, \lambda_t)$ and this good event as $\Xi$.

\begin{proof}
For simplicity, we denote the filtration of history as  $\gH_t := \left( S_1, Y_1, \cdots, S_{t-1}, Y_{t-1}, S_t \right)$. Then we have
\begin{align*}
    \ell_t \sim \exp(\psi(S_t)^\top \theta^*), \quad \E[\ell_t \mid \gH_t] = \frac{1}{\psi(S_t)^\top\theta^*},
\end{align*}
by the property of exponential distribution and definition of $\psi(S)$. Since we have
\begin{align*}
    \hat\theta_t \leftarrow \argmin_{\theta \in \R^d} \gL_t(\theta; \lambda_t),
\end{align*}
by \Cref{alg:k-min}. Then by KKT condition, we have
\begin{align*}
   \left.\frac{\partial \gL_t(\theta ; \lambda_t)}{\partial \theta} \right|_{\theta = \hat\theta_t} = 0 \Rightarrow g_t(\hat\theta_t; \lambda_t) - \sum_{i < t} \ell_i\psi(S_i) = 0
\end{align*}
Notice that by definition of $g_t$,
\begin{align*}
    g_t(\theta^*; \lambda_t) =  \sum_{i < t}  \frac{1}{\psi(S_i)^\top \theta^*} \cdot \psi(S_i) - \lambda_t \theta^*.
\end{align*}
	Denote $\varepsilon_t := \ell_t - \E[\ell_t \mid \gH_t] = \ell_t - {1}/({\psi(S_t)^\top\theta^*})$. We then have
\begin{align*}
    g_t(\hat \theta_t; \lambda_t) - g_t(\theta^*; \lambda_t) = \sum_{i < t} \varepsilon_i \psi(S_i) + \lambda_t \theta^*.
\end{align*}
Fix $s$ such that $|s|\le 1/M_1$. Since $\ell^*(S_t)=1/(\psi(S_t)^\top\theta^*)\le L^*$, this range implies $s<\psi(S_t)^\top\theta^*$, so the moment below is well defined. We have
\begin{align*}
	    \E[\exp(s\varepsilon_t) \mid \gH_t] &= \E\left[\exp\left(s\ell_t - \frac{s}{\psi(S_t)^\top\theta^*}\right)\right] \\
	    &= \exp\left(- \frac{s}{\psi(S_t)^\top\theta^*}\right)\cdot \E\left[\exp(s\ell_t)\mid \gH_t\right],
\end{align*}
and by calculation,
\begin{align*}
    \E[\exp(s\varepsilon_t) \mid \gH_t] &= \exp\left(- \frac{s}{\psi(S_t)^\top\theta^*}\right)\cdot \E\left[\exp(s\ell_t)\mid \gH_t\right] \\
	    &=\exp\left(- \frac{s}{\psi(S_t)^\top\theta^*}\right)\cdot \int_{[0,+\infty)} \psi(S_t)^\top\theta^*\exp(-(\psi(S_t)^\top\theta^* - s)y)dy \\
    &= \exp\left(-\frac{s}{a_t^*} + \log(a_t^*) - \log(a_t^* - s)\right) \\
    &= \exp\left(-x - \log(1-x)\right),
\end{align*}
where $a_t^* := \psi(S_t)^\top\theta^*$ and $x := s/a_t^*$. Since $1/a_t^* = \ell^*(S_t) \le L^*$, the choice $M_1=2L^*$ implies $|x|\le 1/2$ whenever $|s|\le 1/M_1$. For $|x|\le 1/2$, we have $-x-\log(1-x)\le x^2$, and hence
\begin{align*}
    \E[\exp(s\varepsilon_t) \mid \gH_t]
    \le \exp\left(\frac{s^2}{(a_t^*)^2}\right).
\end{align*}
Denote $\nu_{t-1} := 1/{(\psi(S_t)^\top \theta^*)^2}$. Then, for $|s| \le 1/M_1$,
\begin{align*}
    \E[\exp(s\varepsilon_t) \mid \gH_t] \le \exp(s^2 \nu_{t-1}).
\end{align*}
	Applying \citet[Theorem 2]{janz2024exploration} with $\mS_t := \sum_{i<t} \varepsilon_i \psi(S_i)$, we can show that with probability at least $1 - \delta$,
	\begin{align*}
	    \left\|g_t(\hat\theta_t; \lambda_t) - g_t(\theta^*; \lambda_t)\right\|_{H_t^{-1}(\theta^*;\lambda_t)} &\le \left\|\sum_{i<t}\varepsilon_i \psi(S_i)\right\|_{H_t^{-1}(\theta^*;\lambda_t)} + \lambda_t \left\|\theta^*\right\|_{H_t^{-1}(\theta^*;\lambda_t)} \\
    &\le \frac{\sqrt{\lambda_t}}{2M_1} + \frac{2M_1}{\sqrt{\lambda_t}}\log\left(\frac{\det(H_t(\theta^*)^{1/2}/\lambda_t^{d/2})}{\delta}\right) + \frac{2M_1}{\sqrt{\lambda_t}}d\log(2) + \sqrt{\lambda_t} V,
\end{align*}
where $V = \sup\{\|\theta\|_2 : \theta \in \Theta\}$. Moreover, by definition of $H_t(\theta^*; \lambda_t)$, we have
\begin{align*}
    \det(H_t(\theta^*; \lambda_t))/\lambda_t^d \le \left(1 + \frac{tL^*}{\lambda_t d}\right)^d,
\end{align*}

Therefore, for
\begin{align*}
    \gamma_t(\delta, \lambda_t) \ge \sqrt{\lambda_t}\left(\frac{1}{2M_1} + V\right) + \frac{2M_1d}{\sqrt{\lambda_t}}\left(\log(2) + \frac{1}{2}\log\left(1 + \frac{tL^*}{\lambda_t d}\right)\right) + \frac{2M_1}{\sqrt{\lambda_t}}\log(1/\delta),
\end{align*}
we have with probability at least $1 - \delta$,
\begin{align*}
    \theta^* \in C_t(\hat\theta_t; \delta,\lambda_t) := \left\{ \theta \in \Theta : \left\| g_t(\theta; \lambda_t) - g_t(\hat\theta_t; \lambda_t) \right\|_{H_t^{-1}(\theta; \lambda_t)} \le \gamma_t(\delta, \lambda_t) \right\},
\end{align*}
holds for any $t \in [T]$.
\end{proof}
\end{lemma}

\subsection{Proof of Theorem~\ref{thm:kminexp}}
\begin{theorem}[Formal version of \Cref{thm:kminexp}]
\label{thm:kminexp-formal}
By setting $\delta = 1/T$, $\gamma_t(\delta)$ according to \Cref{eq:def-gamma}, and $\lambda_t$ according to \Cref{eq:def-lambda}, \Cref{alg:k-min} enjoys the following regret guarantee:
\begin{align*}
    \gR(T) &\le
    16\gamma \cdot \sqrt{dT}\cdot \sqrt{(\ell^*(S^*))^2(1 + L^*/\lambda) \cdot \log\left(1 + L^*T/d\lambda\right)} \\
    &\quad + 256\gamma^2 \cdot dL^* \cdot \log\left(1 + L^*T/d\lambda\right) \cdot \left(\frac{\sup_{S \in \gS} (\psi(S)^\top \theta^*)}{\ell^*(S^*)^3} + 2\right) + 1,
\end{align*}
	where $\gamma := \sup_t \gamma_t(\delta)$ and $\lambda := \inf_t \lambda_t$.
\begin{proof}
Since we have $X_i \sim \exp(\phi(i)^\top \theta^*)$, then
\begin{align*}
    \min_{i \in S} X_i \sim \exp\left(\sum_{i \in S} \phi(i)^\top \theta^* \right) = \exp\left(\psi(S)^\top \theta^* \right),
\end{align*}
which shows that
\begin{align*}
    \ell^*(S) = \E\left[\min_{i \in S} X_i\right] = \frac{1}{\psi(S)^\top \theta^*}.
\end{align*}
Therefore, by second-order Taylor expansion, we have for some $\xi \in [\ell^*(S_t), \sup_t \ell^*(S_t)]$,
\begin{align*}
    \gR(T) &= \E\left[\sum_{t=1}^T \ell^*(S_t) - \ell^*(S^*) \right]\\
    &\le   \Prob[\Xi] \cdot \E\left[\sum_{t=1}^T \frac{1}{\psi(S_t)^\top \theta^*} - \frac{1}{\psi(S^*)^\top \theta^*} \middle| \Xi \right]+ \Prob[\neg\Xi] \cdot T \\
    &\le \E\left[ \underbrace{\sum_{t=1}^T \frac{1}{(\psi(S_t)^\top \theta^*)^2} \cdot \left( \psi(S^*)^\top \theta^* - \psi(S_t)^\top \theta^*  \right)}_{\gR_1(T)} \middle| \Xi \right] +\E\left[ \underbrace{\sum_{t=1}^T \frac{2}{\xi^3} \cdot \left(\psi(S^*)^\top \theta^* - \psi(S_t)^\top \theta^* \right)^2 }_{\gR_2(T)} \middle| \Xi \right] + 1.
\end{align*}
Under $\Xi$, we have $\theta^* \in C_t(\hat\theta_t; \delta, \lambda_t)$ for every $t \in [T]$. Therefore, by \Cref{alg:k-min}, we have
\begin{align}
\label{eq:optimism}
    \psi(S^*)^\top \theta^* \le \psi(S_t)^\top \wt{\theta}_t.
\end{align}
Under $\Xi$, we have
\begin{align*}
    \gR_1(T) &\le \sum_{t=1}^T \frac{1}{(\psi(S_t)^\top \theta^*)^2} \cdot \psi(S_t)^\top ( \wt{\theta}_t - \theta^*) \\
    &\le \sum_{t=1}^T \frac{1}{(\psi(S_t)^\top \theta^*)^2} \cdot \| \psi(S_t) \|_{H_t^{-1}(\theta^*; \lambda_t)} \cdot \left\|\theta^* - \wt{\theta}_t\right\|_{H_t^{-1}(\theta^*; \lambda_t)},
\end{align*}
where the first inequality is due to \Cref{eq:optimism} and the second holds by Cauchy-Schwartz inequality. Notice that $\wt{\theta}_t, \theta^* \in C_t(\hat\theta_t; \delta, \lambda_t)$ under $\Xi$, we have
\begin{align*}
    \left\|\theta^* - \wt{\theta}_t\right\|_{H_t^{-1}(\theta^*; \lambda_t)} \le 8\gamma_t(\delta, \lambda_t)
\end{align*}
by \citet[Lemma 30]{liu2024almost}. Denote $\gamma := \sup_{t \in [T]} \gamma_t(\delta, \lambda_t)$, we can upper bound $\gR_1(T)$ by
\begin{align*}
    \gR_1(T) \le 8 \cdot \sum_{t=1}^T \frac{1}{(\psi(S_t)^\top \theta^*)^2} \cdot \| \psi(S_t) \|_{H_t^{-1}(\theta^*; \lambda_t)} \cdot \gamma.
\end{align*}
	Denote $A_t := \psi(S_t)/(\psi(S_t)^\top \theta^*)$, we have $H_t(\theta^*; \lambda_t) = \sum_{i < t} A_i A_i^\top + \lambda_t I$ and $\|A_t\|_2 \le \sum_{i \in S_t} \|\phi(i)\|_2 \cdot \ell^*(S_t) \le KL^*$. Then we have
\begin{align*}
    \gR_1(T) &\le 8\gamma \sqrt{\sum_{t=1}^T \|A_t\|^2_{H_t^{-1}(\theta^*; \lambda_t)}} \cdot \sqrt{\sum_{t=1}^T \frac{1}{(\psi(S_t)^\top \theta^*)^2}} \\
    &\le 16\gamma \cdot \sqrt{d(1 + KL^*/\lambda) \cdot \log\left(1 + KL^*T/d\lambda\right)} \cdot \sqrt{\sum_{t=1}^T \frac{1}{(\psi(S_t)^\top \theta^*)^2}},
\end{align*}
where the first inequality is due to the Cauchy-Schwartz inequality, and the second inequality is due to the elliptical potential lemma of \citet{abbasi2011improved}. Moreover, by \citet[Lemma 31]{liu2024almost}, we have
\begin{align*}
	    \sqrt{\sum_{t=1}^T \frac{1}{(\psi(S_t)^\top \theta^*)^2}} &\le \sqrt{T\cdot \frac{1}{(\psi(S^*)^\top\theta^*)^2} + 2 \cdot \gR(T)} \\
    &\le \sqrt{T\cdot (\ell^*(S^*))^2 } + \sqrt{2 \cdot \gR(T)},
\end{align*}
which shows that for $\lambda := \inf_t \lambda_t$,
\begin{align*}
    \gR_1(T) &\le 16\gamma \cdot \sqrt{d(1 + L^*/\lambda) \cdot \log\left(1 + L^*T/d\lambda\right)} \cdot \sqrt{T\cdot (\ell^*(S^*))^2 } \\
    &\quad + 16\gamma \cdot \sqrt{d(1 + L^*/\lambda) \cdot \log\left(1 + L^*T/d\lambda\right)} \cdot \sqrt{2 \cdot \gR(T)}.
\end{align*}
Next we give the upper bound for $\gR_2(T)$. Recall that
\begin{align*}
	    \gR_2(T) = \sum_{t=1}^T \frac{2}{\xi^3} \cdot \left(\psi(S_t)^\top \theta^* - \psi(S^*)^\top\theta^*\right)^2.
\end{align*}
Then, under $\Xi$, we have
\begin{align*}
    \gR_2(T) &\le \sum_{t=1}^T \frac{2}{\xi^3} \cdot \left\langle\psi(S_t) ,\theta^* - \wt{\theta}_t\right\rangle^2 \\
    &\le  \frac{2}{\ell^*(S^*)^3} \cdot \sum_{t=1}^T \|\psi(S_t)\|^2_{H_t^{-1}(\theta^*; \lambda_t)} \cdot \|\theta^* - \wt{\theta}_t\|^2_{H_t^{-1}(\theta^*; \lambda_t)} \\
    &\le \frac{2}{\ell^*(S^*)^3} \cdot 64\gamma^2 \cdot \sum_{t=1}^T \|\psi(S_t)\|^2_{H_t^{-1}(\theta^*; \lambda_t)},
\end{align*}
where the first inequality is according to \Cref{eq:optimism}, the second inequality is due to the Cauchy-Schwartz inequality, and the last inequality holds by \Cref{lemma:mle-concentration}. Denote
\begin{align*}
	    \Lambda_t := \lambda_t I + \sum_{i < t} \psi(S_i)\psi(S_i)^\top.
\end{align*}
Then we have
\begin{align*}
    \sup_{S \in \gS} (\psi(S)^\top \theta^*) \cdot \Lambda_t^{-1} \succ H_t^{-1}(\theta^*; \lambda_t),
\end{align*}
which further implies
\begin{align*}
    \gR_2(T) &\le \frac{2}{\ell^*(S^*)^3} \cdot 64\gamma^2 \cdot \sup_{S \in \gS} (\psi(S)^\top \theta^*) \cdot \sum_{t=1}^T \|\psi(S_t)\|^2_{\Lambda_t^{-1}} \\
    &\le \frac{2}{\ell^*(S^*)^3} \cdot 64\gamma^2 \cdot \sup_{S \in \gS} (\psi(S)^\top \theta^*) \cdot 2dL^*\log(1 + L^* T/d\lambda) \\
    &= \frac{256}{\ell^*(S^*)^3} \sup_{S \in \gS} (\psi(S)^\top \theta^*) \cdot \gamma^2 \cdot dL^*\log(1 + L^*T/d\lambda).
\end{align*}
Therefore, we have
\begin{align*}
    \gR(T) &\le 16\gamma \cdot \sqrt{d(1 + L^*/\lambda) \cdot \log\left(1 + L^*T/d\lambda\right)} \cdot \sqrt{T\cdot (\ell^*(S^*))^2 } \\
    &\quad + 16\gamma \cdot \sqrt{d(1 + L^*/\lambda) \cdot \log\left(1 + L^*T/d\lambda\right)} \cdot \sqrt{2 \cdot \gR(T)} \\
    &\quad + \frac{256}{\ell^*(S^*)^3} \sup_{S \in \gS} (\psi(S)^\top \theta^*) \cdot \gamma^2 \cdot dL^*\log(1 + L^*T/d\lambda) + 1.
\end{align*}
Notice that for $x \le A\sqrt{x} + B$, we have $x \le 2A^2 + B$. Therefore, we have
\begin{align*}
    \gR(T) &\le
    16\gamma \cdot \sqrt{dT}\cdot \sqrt{(\ell^*(S^*))^2(1 + L^*/\lambda) \cdot \log\left(1 + L^*T/d\lambda\right)} \\
    &\quad + 256\gamma^2 \cdot dL^* \cdot \log\left(1 + L^*T/d\lambda\right) \cdot \left(\frac{\sup_{S \in \gS} (\psi(S)^\top \theta^*)}{\ell^*(S^*)^3} + 2\right) + 1, \\
    &\le \wt{\gO}\left(\sqrt{d^3 T}\right)
\end{align*}

\end{proof}
\end{theorem}

\section{Regret Lower Bound for K-Min Exponential Bandits}
\label{app:lowerbound}

For completeness, we provide the regret lower bound for $K$-Min Exponential Bandits under full-bandit feedback. This lower bound, $\Omega(\sqrt{T})$, illustrates that \texttt{MLE-EXP} is nearly optimal, matching the regret upper bound $\wt{O}(\sqrt{T})$. The construction and proof of the lower bound are standard.

\begin{theorem}[Regret lower bound for $K$-Min Exponential Bandits]
Consider the $K$-Min Exponential Bandits defined in \Cref{sec:kminexp} with $N \ge K+1$ arms. There exists a constant $c>0$ and a problem instance such that any learning algorithm satisfies
\[
\gR(T) \;\ge\; c \sqrt{T}.
\]
\end{theorem}

\begin{proof}
We construct the hard-to-learn instance with two close actions and apply the standard method in \citet{lattimore2020bandit} to establish the lower bound.

Fix $m>0$ and a small parameter $\Delta \in (0,m)$.
Construct $K-1$ base arms $B=\{1,\dots,K-1\}$ each with rate $\mu_i = m, i \in B$, and two ``contender'' arms $a$ and $b$ whose rates differ across two environments:
\[
\begin{aligned}
\mathsf{P}: &\quad \mu_a = m+\Delta,\;\; \mu_b = m,\\
\mathsf{Q}: &\quad \mu_a = m,\;\; \mu_b = m+\Delta.
\end{aligned}
\]
Any valid action must select all arms in $B$ and exactly one of $\{a,b\}$.
Thus there are two actions: $S^a = B\cup\{a\}$ and $S^b = B\cup\{b\}$.
Under $\mathsf{P}$, $S^a$ is optimal; under $\mathsf{Q}$, $S^b$ is optimal.

Notice that $\min_{i\in S}X_i \sim \Exp(\sum_{i\in S}\mu_i)$. Hence
\[
\mu_{S^a} = \sum_{i \in S^a} \mu_i = K m + \Delta,\qquad \mu_{S^b} = \sum_{i \in S^b} \mu_i = K m.
\]
The one-step regret gap is
\[
\Delta_{\mathrm{gap}}
= \ell^\star(S^b) - \ell^\star(S^a)
= \frac{1}{K m} - \frac{1}{K m+\Delta}
= \frac{\Delta}{(K m)(K m+\Delta)}
\asymp \frac{\Delta}{(K m)^2}.
\]

Next, consider the information distance.
For exponential laws,
\[
\KL\!\big(\Exp(\mu)\;\|\;\Exp(\mu')\big)
= \log\frac{\mu}{\mu'} -1 + \frac{\mu'}{\mu}.
\]
Setting $\mu=K m+\Delta$, $\mu'=K m$, the divergence expands to $\tfrac12 (\Delta/(K m))^2+O((\Delta/(K m))^3)$.
Thus each play contributes $O((\Delta/(K m))^2)$ to the total KL.
Over $T$ rounds,
\[
\KL(\mathsf{P}\;\|\;\mathsf{Q}) \;\le\; c_0 T \Big(\tfrac{\Delta}{K m}\Big)^2
\]
for some $c_0>0$.

By the Bretagnolle–Huber inequality,
\[
\frac{\gR_{\mathsf{P}}(T) + \gR_{\mathsf{Q}}(T)}{2}
\;\ge\; \Delta_{\mathrm{gap}} \cdot \frac{T}{2} \exp\!\Big(-\KL(\mathsf{P}\;\|\;\mathsf{Q})\Big).
\]

Finally, choose $\Delta = \alpha K m/\sqrt{T}$ for a small constant $\alpha>0$.
Then $\KL(\mathsf{P}\;\|\;\mathsf{Q})=O(1)$ and $\Delta_{\mathrm{gap}}\asymp (\alpha/(K m)) \cdot 1/\sqrt{T}$.
Therefore
\[
\frac{\gR_{\mathsf{P}}(T) + \gR_{\mathsf{Q}}(T)}{2} \;\ge\; c_1 \sqrt{T},
\]
for some $c_1>0$. Hence in at least one environment the regret satisfies $\gR(T)\ge c\sqrt{T}$, completing the proof.
\end{proof}

\end{document}